\documentclass[conference]{IEEEtran}
\IEEEoverridecommandlockouts
\usepackage{times}

\usepackage[numbers]{natbib}
\usepackage{multicol}
\usepackage[bookmarks=true]{hyperref}

\let\labelindent\relax
\usepackage{url}
\usepackage{booktabs}
\usepackage{graphicx}
\usepackage{subcaption} 
\usepackage{wrapfig}
\usepackage{amsmath}
\usepackage{amssymb}
\usepackage{enumitem}
\usepackage{xspace}
\usepackage{comment}
\usepackage{algorithm}
\usepackage{algpseudocode}
\usepackage{longtable} 
\usepackage{multirow}
\usepackage{xcolor}
\usepackage{tabularx}

\setlist[itemize]{leftmargin=*}
\setlist[enumerate]{leftmargin=*}

\newcommand{\method}{{\it DISC}\xspace}

\begin{document}

\title{DISC: Decoupling Instruction from State-Conditioned Control via Policy Generation}
\author{
  \IEEEauthorblockN{
    Hanxiang Ren\textsuperscript{\ddag,\S,\P,*} \quad
    Pei Zhou\textsuperscript{\S,\P,*} \quad
    Xunzhe Zhou\textsuperscript{\S,\P} \quad
    Yanchao Yang\textsuperscript{\S,\P,\dag}
  }
  \IEEEauthorblockA{
    \textsuperscript{\ddag}Zhejiang University \quad
    \textsuperscript{\S}The University of Hong Kong \quad
    \textsuperscript{\P}TranscEngram
  }
  \thanks{\textsuperscript{*}Equal contribution.\ \textsuperscript{\dag}Corresponding author.}
}
\maketitle

\begin{abstract}
Language-conditioned manipulation policies 
typically process instructions and observations 
through shared network parameters.
This \emph{task-state entanglement} 
provides a pathway
for \emph{observation leakage} --
networks learn scene-to-action shortcuts
that bypass language grounding entirely.
\method eliminates this failure structurally.
Rather than conditioning a universal policy on language,
\method uses a hypernetwork to generate
the \emph{entire parameter set} of a task-specific visuomotor policy
from the instruction alone.
The generated policy never directly accesses language;
therefore, its task-awareness must come from the language.
Consequently,
observation leakage has no pathway to emerge.
On the other hand, 
generating coherent high-dimensional policy weights
is itself a challenging problem.
We address it with a two-stage hypernetwork
whose refinement stage embeds the structure
of gradient-based optimization as a feed-forward inductive bias,
producing globally consistent parameters
without actual gradient computation.
Trained entirely from scratch on standard data budgets,
\method outperforms all entangled baselines
on LIBERO-90 and Meta-World,
with advantages that widen on complex, long-horizon tasks --
and surpasses the large-scale pretrained $\pi_0$
despite using no external pretraining data.
On a real-world benchmark
where all tasks share identical visual context,
\method substantially outperforms entangled alternatives,
directly confirming that language-generated policy parameters,
not visual shortcuts, drive behavior.
The hypernetwork further learns a semantically structured
parameter manifold that enables
few-shot adaptation from minimal demonstrations
and robust generalization across paraphrased instructions.
Our code is available at: \href{https://github.com/ReNginx/DISC}{https://github.com/ReNginx/DISC}.
\end{abstract}

\IEEEpeerreviewmaketitle

\section{Introduction}

Reliable task-conditioned manipulation requires robots to treat the task specification as the source of task identity, 
rather than as a weak hint that can be overridden by familiar visual scenes. 
Language-conditioned manipulation policies 
must process two fundamentally different inputs: 
an instruction that specifies \emph{what task to perform} (fixed throughout an episode) 
and an observation that captures \emph{what the current state is} (evolving continuously). 
Current architectures 
conflate these roles by processing both through 
shared representations -- 
a design we term \emph{task-state entanglement}. 
We argue this creates two compounding failures. 
{\it First,} 
shared parameters 
face optimization tension 
between language grounding 
(requiring semantic understanding) 
and visuomotor control 
(requiring spatial precision). 
{\it Second,} 
and more critically, 
entangled architectures enable \emph{observation leakage}: 
visual context often correlates with task identity 
in limited robotics datasets, 
allowing the network 
to map scenes 
directly to actions while bypassing language entirely. 
For example, in a shared kitchen scene, 
a policy may learn the frequent pan-placement behavior 
and omit the instructed stove-activation subgoal. 
The result is policies that exploit visual shortcuts 
rather than ground instructions -- 
succeeding on familiar scenes but failing when 
instructions change or visual context becomes ambiguous. 
Evidence from the broader vision-language community 
confirms this pattern, 
manifesting as modality competition~\citep{liu2025seeing,tang2025shaping}, 
visual neglect~\citep{mullick2025text}, 
and entity grounding failures~\citep{alonso2025vision,pani2025gaze}.

We propose \method, 
which takes architectural decoupling to its logical conclusion: 
a task specification, instantiated here as language, generates the \emph{entire parameter set} of a task-specific visuomotor policy. 
A hypernetwork processes instructions alone to produce policy weights; the generated policy processes visual observations alone using those weights. 
This separation provides a structural guarantee against observation leakage -- no task-agnostic parameters exist where shortcuts could emerge, and task semantics influence behavior only through generated weights.
A central challenge is that naive weight generation produces incoherent parameters. \method addresses this with a two-stage architecture. 
A Weight Initialization Network first maps the language embedding to a semantically informed point in parameter space. 
An Iterative Refinement Module then improves this initialization by mimicking the structure of gradient-based optimization -- forward evaluation, error estimation, and parameter correction -- while remaining fully feed-forward at inference time. 
This optimization-inspired inductive bias enables the hypernetwork to produce globally coherent policy parameters without incurring the cost of actual gradient computation, bridging the gap between the expressiveness of full policy generation and the quality demands of precise visuomotor control.

Despite training entirely from scratch, 
\method achieves 94.3\% on LIBERO-90 and 92.2\% on Meta-World, outperforming the strongest trained-from-scratch entangled baseline by 7.7\% on LIBERO-90. 
For context, \method surpasses even the pretrained $\pi_0$ (91.6\%) and remains competitive with $\pi_{0.5}$ (95.7\%) -- demonstrating that architectural inductive biases 
can recover much of the benefit attributed 
to large-scale pretraining. 
The advantage widens on complex, 
long-horizon tasks, precisely where task-state entanglement is most damaging. 
On a real-world combinatorial benchmark 
where visual context is shared across all 9 tasks, 
\method achieves 86.4\% versus 78.5\% for the best entangled baseline, confirming that task-specific generated parameters, 
not visual shortcuts, drive behavior. 
The decoupled architecture further enables superior few-shot adaptation, robust grounding across paraphrased instructions, and a semantically structured parameter manifold where related tasks cluster closely by functional meaning.

\section{Related Work}
\label{sec:related_work}

\textbf{Multi-Task Manipulation Policies.}
Current language-conditioned manipulation methods 
process instructions and observations 
through shared network components 
and differ mainly in their fusion mechanism.
Transformer-based approaches~\citep{bu2025univla,kim2024openvla,zhang2025dreamvla,zitkovich2023rt,zhou2024maxmi} 
tokenize and concatenate multimodal inputs 
into unified sequences~\citep{vaswani2017attention}, 
as exemplified by RT-1~\citep{brohan2022rt}, Gato~\citep{reed2022generalist}, 
Octo~\citep{team2024octo}, 
VQ-BeT~\citep{shafiullah2022behavior}, and BAKU~\citep{haldar2024baku} -- 
where the same $W_Q$, $W_K$, $W_V$ projections and feed-forward weights 
operate on both language and visual tokens.
Text-aware variants such as OTTER~\citep{huang2025otter} condition visual features on language, but retain shared action-prediction parameters.
Concept-guided policies such as AutoCGP~\citep{zhou2025autocgp} learn semantic control structure from unlabeled demonstrations.
Diffusion-based methods~\citep{ho2020denoising,song2020denoising} 
instead cast control as conditional denoising, pioneered by Diffusion Policy~\citep{chi2023diffusion}.
Architectures have evolved from U-Nets~\citep{ronneberger2015u} with FiLM conditioning~\citep{perez2018film} to Transformer-based backbones such as DiT~\citep{peebles2023scalable} and BESO~\citep{reuss2023goal}, 
with RDT-1B~\citep{liu2024rdt} and MoDE~\citep{reuss2024efficient} 
scaling through pre-training and mixture-of-experts tuning, while DP3~\citep{ze20243d}, 3D Diffuser Actor~\citep{ke20243d}, and PointVLA~\citep{li2026pointvla} enhance geometric grounding via 3D representations.
Flow-matching models $\pi_0$~\citep{black2024pi_0} and $\pi_{0.5}$~\citep{intelligence2025pi_} further improve generalization through large-scale pre-training.
Despite this diversity, all these methods entangle instruction and state within shared processing stages.
Concatenation-based transformers apply identical 
weights to both modalities throughout the network.
Cross-attention and FiLM-based architectures~\citep{perez2018film} introduce partial separation -- FiLM affects vision only through affine modulation -- but language still controls only a small fraction of computation, 
while the vast majority of parameters remain task-agnostic and can learn shortcut mappings that bypass language entirely.
As analyzed in Section~\ref{sec:limit_task_state}, 
this task-state entanglement creates two problems: optimization tension between competing functions within shared parameters, and observation leakage that enables the network to map visual context directly to actions without grounding language.

\textbf{Hypernetworks for Policy Generation.}
An alternative paradigm 
uses hypernetworks~\citep{beck2023hypernetworks,ha2016hypernetworks,huang2021continual,zhou2026textit} -- 
networks trained to output the weights of another -- 
to generate task-specific policies.
HyPoGen~\citep{ren2025hypogen}, HyperZero~\citep{rezaei2023hypernetworks}, and Hyper-GoalNet~\citep{zhou2026textit} demonstrate this for multi-task or goal-conditioned manipulation.
These methods motivate separating task specification from policy execution, 
but differ in conditioning and generator design: HyPoGen and HyperZero use compact MLP-style generators, while Hyper-GoalNet conditions on goal images and current observations rather than language-only specifications.
\method addresses both the framework question -- 
decoupling instruction from state-conditioned control by generating task-specific target policy parameters from language -- 
and the architecture question, 
introducing an optimization-inspired hypernetwork whose iterative refinement enables feed-forward generation of high-quality policy parameters.
Although adaptable to meta-learning~\citep{finn2017model,santoro2016meta,snell2017prototypical}, we focus on multi-task manipulation policy learning.

\begin{figure*}[t!]
    \centering
    \includegraphics[width=\linewidth]{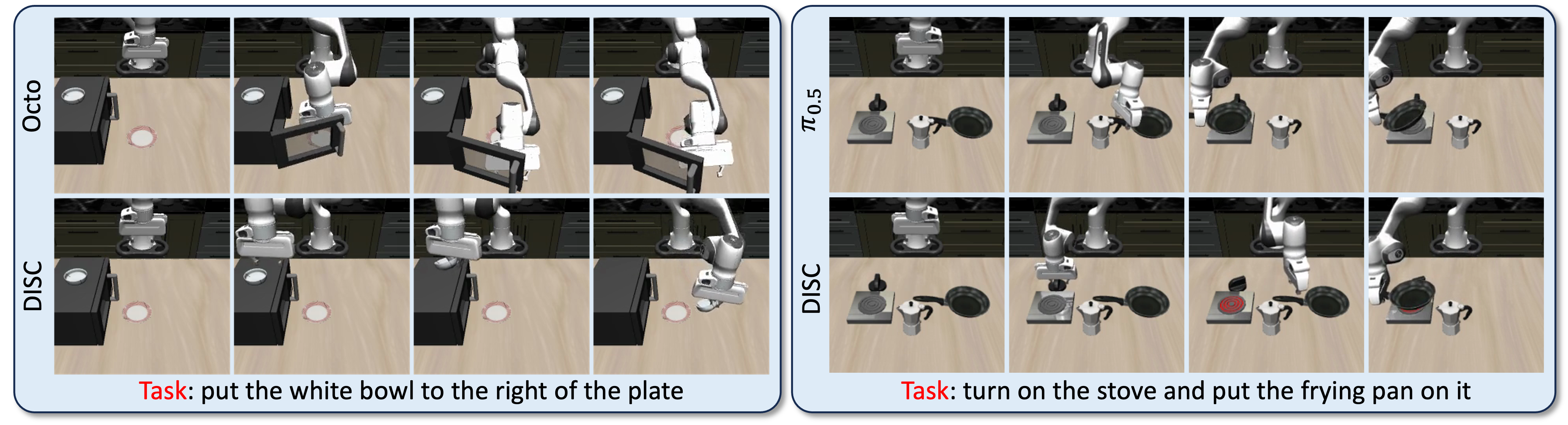}
    \caption{
     \textbf{Behavioral Evidence of Task-State Entanglement.} 
    Current policies trained with entangled architectures fail to ground language instructions faithfully. 
    Left: given the instruction regarding ``white bowl,'' 
    Octo instead approaches the microwave -- executing a behavior associated with a different task that shares similar visual context. 
    Right: given the multi-step instruction ``turn on the stove and put the frying pan on it,'' 
    the large-scale pretrained $\pi_{0.5}$ skips the stove-activation subgoal and directly places the pan on the stove, 
    matching a related pan-placement task under the same scene-level context. 
    These failures illustrate observation leakage: 
    the policies have learned scene-to-action mappings 
    rather than instruction-to-action mappings. 
    \method, with its decoupled architecture, correctly executes both instructions by separating task specification from state-conditioned control.
    }
    \label{fig:attention_map_main}
\end{figure*}

\section{Preliminary}
\label{sec:prelim}
We target multi-task robotic manipulation, where an agent must execute diverse tasks specified by natural language instructions. 
Each task is defined by an instruction 
$l \in \mathcal{J}$, 
and at each timestep, 
the agent receives an observation 
$o_t = (I_t, s_t)$ comprising an RGB image and proprioceptive state, 
from which it produces 
a continuous action $a_t \in \mathcal{A}$.

Notably, 
the instruction and observation 
serve qualitatively different roles. 
The instruction specifies 
\emph{what task to perform} -- 
it defines the goal and remains fixed throughout an episode. 
The observation captures 
\emph{what the current state is} -- 
it evolves continuously and directly constrains which action is appropriate. 
A well-designed policy 
shall treat these inputs 
according to their distinct roles, 
without conflating the two.

For learning,
given a dataset $\mathcal{D}$ of expert demonstrations spanning $N$ tasks, 
we train a policy via behavior cloning:
\begin{equation}
\mathcal{L}_{\text{BC}}(\theta) = \mathbb{E}_{(o_t, a_t, l) \sim \mathcal{D}} \left[ \| \pi(o_t, l; \theta) - a_t \|_2^2 \right] \,.
\label{eq:bc_loss}
\end{equation}

Current approaches learn an entangled policy $\pi(o, l; \theta)$ that processes 
both inputs through shared parameters -- 
a design we term \textbf{task-state entanglement}. 
As we argue next, 
this architectural choice introduces fundamental learning difficulties 
that compromise language grounding.

\subsection{Limitations of Task-State Entanglement}
\label{sec:limit_task_state}

Task-state entanglement 
creates two distinct problems 
that undermine robust instruction following.

\paragraph{Problem 1: Competing Functions within Shared Parameters}
A language-conditioned policy 
must serve two distinct roles: 
interpreting the instruction to understand what task to perform, 
and processing observations to predict appropriate actions. 
In entangled architectures, 
both roles are handled by the same parameters. 
For instance, transformer-based policies concatenate language and visual tokens:
\begin{equation}
\text{Output} = \text{Transformer}(\text{Concat}(T_l, T_v); \theta)\,;
\end{equation}
Diffusion policies similarly condition on jointly embeddings:
\begin{equation}
a_t = \text{Denoise}(\epsilon_t \mid e_l, e_v; \theta)\,.
\end{equation}
In both designs, shared parameters $\theta$ must simultaneously learn language understanding and visuomotor control. 
However, demands conflict: language understanding benefits from semantic abstraction and visual invariance, 
while visuomotor control requires sensitivity to precise spatial details. 
Thus, forcing both roles through shared parameters creates optimization tension that degrades either core function.

\paragraph{Problem 2: Observation Leakage Enables Shortcut Learning}
A more severe failure arises from what we call \textbf{observation leakage}: 
visual observations often contain cues correlating with task identity. 
Specific objects, scene layouts, or robot configurations may appear predominantly in certain tasks. 
Given this,
entangled architectures can learn to map visual patterns directly to actions -- 
bypassing language. 
This shortcut minimizes training loss 
without requiring the network to ground linguistic meaning, 
yielding policies that succeed on familiar scenes but fail when instructions change.

Figure~\ref{fig:attention_map_main} demonstrates this failure. 
In the left example, 
Octo approaches the microwave when asked to put the white bowl to the right of the plate, 
executing a behavior linked to a similar visual context. 
In the right example, 
the pretrained $\pi_{0.5}$ is asked to turn on the stove and put the frying pan on it, 
but it performs only the pan-placement subgoal, 
consistent with a scene-level shortcut from the related task of putting the pan on the stove under the same scene context. 
These examples show that observation leakage can persist across model scales when language and state are fused during action fine-tuning. 
In contrast, our decoupled architecture executes the specified subgoals, 
confirming that architectural separation prevents such shortcuts.
These problems compound to produce policies that are brittle to instruction variation, fail to generalize across visual contexts, and degrade on long-horizon tasks where grounding errors accumulate. 
To overcome these limitations, 
we propose treating language not as an input to be fused, 
but as a generative signal that produces the parameters of a task-specific visuomotor policy -- an approach we detail next.

\begin{figure*}[t!]
    \centering
    \includegraphics[width=\linewidth]{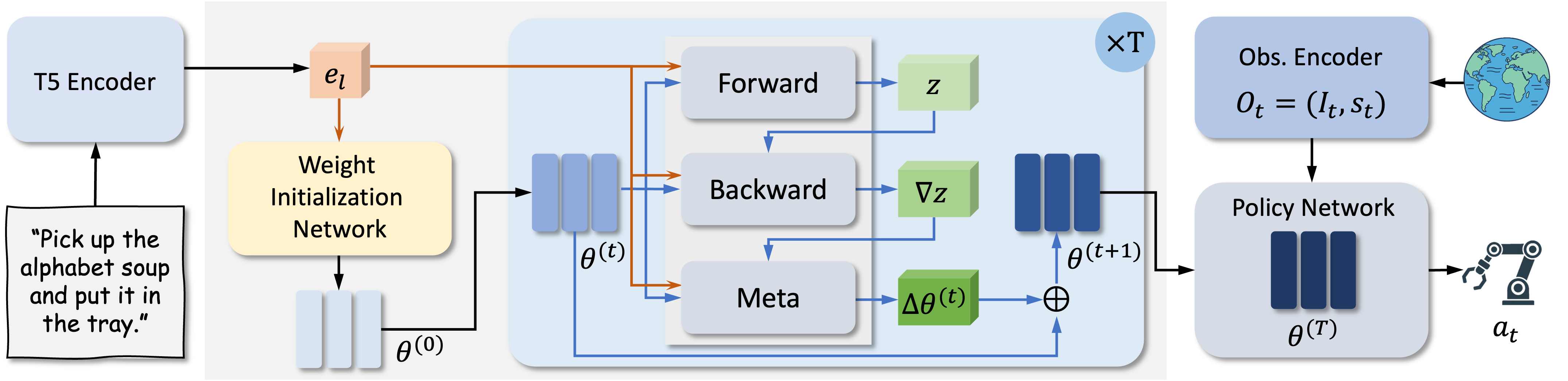}
    \caption{
    \textbf{The \method architecture.} 
    The task specification, instantiated as a language instruction $l$, 
    is encoded into embedding $e_l$ and processed by the hypernetwork through two stages: 
    (1) the Weight Initialization Network generates initial parameters $\theta_\pi^{(0)}$ (labeled as $\theta^{(0)}$), and 
    (2) the learned iterative refinement module updates parameters over $T$ steps to produce final $\theta^{(T)}$. 
    The refinement module mimics optimization structure -- 
    forward, backward, and meta-update operations -- but all transformations are learned, and the entire generation process is feed-forward. 
    The generated parameters configure the target policy $\pi_{\theta_\pi}$, a compact, task-specific visuomotor controller. 
    Observations $o_t$ enter only at the target policy stage: 
    the task specification determines \emph{which} policy executes, while the generated policy processes visual input independently, 
    ensuring complete separation between task-specification processing and state-conditioned control.}
    \label{fig:architecture_overview}
\end{figure*}

\section{Method}
\label{sec:method}

\subsection{Decoupling Instruction from State-Conditioned Control}
\label{subsec:method}

Our solution is to separate 
task-specification processing from state-conditioned control entirely. 
In this paper, 
the task specification is instantiated as a natural-language instruction.
Rather than learning a single policy 
that fuses both modalities, 
our method, \method, 
decomposes the problem 
into two stages:
(1) a hypernetwork $\mathcal{H}_\phi$ 
processes the task specification to generate a complete set of policy weights, 
and 
(2) the generated policy $\pi_{\theta_\pi}$ maps observations to actions 
using only those weights -- without any direct task-specification input.

Formally, 
given an instruction $l$ as the task specification, 
we first obtain a task embedding via a frozen language encoder:
\begin{equation}
e_l = \Phi_L(l) \,.
\label{eq:lang_encoding}
\end{equation}
The hypernetwork then generates policy parameters:
\begin{equation}
\theta_\pi = \mathcal{H}_\phi(e_l)\,.
\label{eq:param_generation}
\end{equation}
Further, 
the generated policy performs visuomotor control:
\begin{equation}
a_t = \pi(o_t; \theta_\pi)\,.
\label{eq:target_policy}
\end{equation}

This decomposition 
directly addresses both problems 
identified in Section~\ref{sec:limit_task_state}. 
{\it First,} 
the competing functions are assigned to separate networks: 
the hypernetwork learns to interpret the task specification, while the target policy learns 
to map observation embeddings to actions, 
eliminating optimization tension between task understanding and visuomotor control. 
{\it Second,} 
the direct action-mapping pathway for observation leakage is structurally removed. 
In entangled architectures, 
shortcuts arise because the model can learn to ignore the task specification and map observations directly to actions -- 
both modalities feed into the same network, 
so bypassing one is easy. 
In \method, 
the target policy has no direct access to the instruction at all. 
The only way for task information to influence behavior is through the generated parameters; 
the hypernetwork \emph{must} encode task semantics into the weights because 
there is no alternative route.

Figure~\ref{fig:architecture_overview} illustrates this architecture. 
Observations $o_t$ enter only at the target policy stage, maintaining complete separation from task-specification processing.

\paragraph{Contrast with Conditioning Approaches}
Existing conditioning methods -- 
concatenation~\cite{shafiullah2022behavior,team2024octo}, 
cross-attention~\cite{li2023vision}, 
and FiLM~\cite{chi2023diffusion,perez2018film} -- 
integrate language and vision within shared processing stages, 
to different extents.

\textit{Concatenation} exhibits the most direct form of entanglement: language tokens $T_l$ and visual tokens $T_v$ are concatenated into a unified sequence and processed by the same transformer layers.
\textit{Cross-attention} uses modality-specific projections to compute queries from vision and keys/values from language.
While these projections are separate, 
the attention operation produces fused features that are processed by subsequent shared layers, entangling the two modalities from that point onward.
\textit{FiLM} is closest in spirit to our approach: language affects vision through learned affine modulation ($\gamma(e_l) \cdot h_v + \beta(e_l)$), and no layer processes both modalities with shared parameters.
{\it However,} 
language generates only a small fraction of the computation -- 
the scale and shift values -- 
while the vast majority of the visual backbone parameters remain shared across all tasks.
These shared, task-agnostic parameters 
can still learn shortcut mappings from visual features to actions that bypass language conditioning entirely.

\method takes this principle to its logical conclusion: 
\emph{all} task-specific parameters of the target policy 
are generated from the task specification, instantiated here as a language instruction, leaving no space in the target policy 
where shortcuts could emerge.
The policy's entire computational structure 
is determined by the task specification, 
ensuring that task semantics are embedded throughout the visuomotor controller rather than applied as a light modulation.

\subsection{Hypernetwork Architecture}
\label{subsec:hypernetwork}

A core challenge is generating coherent, high-performing policy parameters from a task embedding induced by a language instruction. 
Naive direct regression from $e_l$ to the full parameter vector $\theta_\pi$ fails because policy networks contain 
millions of interdependent parameters with complex functional relationships. 
Prior hypernetwork architectures 
(e.g., simple MLP generators) 
struggle to capture these dependencies, 
producing weights that lack the coherent structure required for effective control.

Our key architectural contribution 
is a hypernetwork that incorporates the \emph{structural inductive bias of iterative optimization} while remaining \emph{fully feed-forward at inference time}. 
The design is inspired by a simple observation: gradient-based optimization naturally produces coherent parameters via incremental updates respecting inter-layer dependencies. 
We embed this structure into the architecture itself -- 
the hypernetwork \emph{mimics} the computational pattern of optimization (forward pass, backward pass, parameter update), 
but \emph{how} each of these operations refines the parameters is entirely learned from data. 
This achieves the coherence benefits of optimization-like refinement without incurring the computational cost of actual gradient computation at inference time.

\paragraph{Stage 1: Weight Initialization Network}
The first stage generates a coarse initialization from the task embedding:
\begin{equation}
\theta_\pi^{(0)} = \text{WIN}_{\phi_1}(e_l)\,.
\label{eq:win}
\end{equation}
This positions the parameters 
in a well-conditioned region of the weight space -- 
distinguishing, for instance, 
grasping tasks from pushing tasks -- 
but lacks the fine-grained precision required for successful execution.

\paragraph{Stage 2: Learned Iterative Refinement}
The second stage refines this initialization through $T$ learned update steps:
\begin{equation}
\theta_\pi^{(t+1)} = \theta_\pi^{(t)} + \Delta\theta^{(t)}, \quad \text{where} \quad \Delta\theta^{(t)} = f_{\phi_2}(\theta_\pi^{(t)}, e_l)\,.
\label{eq:refinement}
\end{equation}

Importantly, 
this is \emph{not} optimization in the traditional sense. 
The refinement module $f_{\phi_2}$ is a neural network with fixed, learned parameters $\phi_2$. 
At inference time, 
generating policy parameters requires only $T$ forward passes through $f_{\phi_2}$ -- 
no gradients are computed with respect to any loss, 
no demonstration data is accessed, and no iterative solving occurs. 
The entire generation process is feed-forward.

Specifically, 
the refinement module is architecturally decomposed to mirror the structure of optimization: 
it computes a latent forward representation 
$z$ from the current parameters, 
estimates pseudo-gradient signals $\nabla z$, and aggregates these into parameter updates $\Delta\theta^{(t)}$ (see Figure~\ref{fig:architecture_overview}). 
However, unlike true optimization where gradients are computed analytically from a loss function, 
here the ``forward,'' ``backward,'' and ``update'' operations are all learned neural network components. 
The architecture provides the structural scaffold -- the pattern of how information flows -- 
while training determines the actual transformation at each stage. 
This learned refinement process produces globally coherent parameter updates, 
similar to how true gradients enforce consistency across layers, 
but adapted specifically to the encountered task distribution.

\paragraph{Target Policy}
The generated parameters $\theta_\pi$ configure a compact visuomotor controller mapping observations to actions. 
Because the target policy is specialized to a single task, 
it can be significantly smaller than monolithic architectures that must accommodate the entire task distribution within one set of weights. 
In our implementation, 
the target policy is a lightweight MLP, enabling efficient inference suitable 
for high-frequency robotic control. 
More architectural details 
are provided in the appendix.

\paragraph{Training and Inference}
The full framework is trained end-to-end by optimizing hypernetwork parameters $\phi = \{\phi_1, \phi_2\}$ using the behavior cloning loss:
\begin{equation}
\mathcal{L}_{\text{BC}}(\phi) = \mathbb{E}_{(o_t, a_t, l) \sim \mathcal{D}} \left[ \| \pi(o_t; \mathcal{H}_\phi(\Phi_L(l))) - a_t \|_2^2 \right]\,.
\label{eq:bc_loss_full}
\end{equation}
For each training sample, 
the hypernetwork generates policy parameters $\theta_\pi = \theta_\pi^{(T)}$ from the task specification, 
the target policy uses these parameters to predict actions, 
and the prediction error backpropagates through the entire generation process to update $\phi$ accordingly.
At inference time, 
deployment is straightforward: given a new instruction $l$ as the task specification, we encode it, run the hypernetwork forward to generate $\theta_\pi$, and execute the target policy in the environment. 
The generation is a single feed-forward pass through the hypernetwork for each task, 
after which the compact target policy runs independently at control frequency without further overhead during execution.

\subsection{Few-Shot Adaptation (Optional)}
\label{subsec:adaptation}

We call this adaptation optional because it is not required by the standard inference pipeline; it is used only when a novel task is accompanied by a few demonstrations.
A practical benefit 
of \method's decoupled architecture 
is efficient adaptation 
to novel tasks from minimal demonstrations. 
The separation between 
the hypernetwork (which captures general knowledge about how language maps to policy structure) 
and the target policy (which executes a specific task) provides a natural decomposition for efficient adaptation.

When presented 
with a novel task 
with instruction $l_{\text{new}}$ 
and a small number of demonstrations $\mathcal{D}_{\text{new}} = \{\xi_j\}_{j=1}^K$, 
adaptation proceeds in two stages.
{\it First,} 
the hypernetwork 
generates by mapping the new instruction to policy parameters:
\begin{equation}
\theta_\pi^{\text{init}} = \mathcal{H}_\phi(\Phi_L(l_{\text{new}}))\,.
\label{eq:init_adaptation}
\end{equation}
Since the hypernetwork 
has learned the structure of the task manifold during multi-task training, 
this initialization already encodes 
relevant task semantics -- 
positioning the parameters near the task-optimal region 
rather than at a random starting point.
{\it Second,} 
the generated target policy parameters 
are fine-tuned 
with the provided demonstrations:
\begin{equation}
\theta_\pi^* = \arg\min_{\theta_\pi} \sum_{\xi_j \in \mathcal{D}_{\text{new}}} \sum_{(o_t, a_t) \sim \xi_j} \left\| \pi(o_t; \theta_\pi) - a_t \right\|_2^2\,.
\label{eq:adaptation_loss}
\end{equation}
The hypernetwork 
remains frozen throughout, 
preserving its learned meta-knowledge. 
This transforms adaptation from 
a global search over a large, entangled parameter space into a local refinement of a compact, 
well-initialized policy -- 
requiring fewer demonstrations and fewer gradient steps to reach competent performance.

\section{Experiment}
\label{sec:exp}

We conduct experiments 
to validate that explicit decoupling 
of instruction processing from state-conditioned control 
through \method improves language-conditioned robotic manipulation. 
Our evaluation addresses 
four key research questions:
\begin{enumerate}
    \item \textbf{Multi-Task Efficiency (RQ1):} 
    Does \method's decoupled architecture 
    outperform state-of-the-art entangled methods 
    on diverse manipulation tasks?    
    \item \textbf{Few-Shot Adaptation (RQ2):} 
    Can the structured parameter manifold learned 
    by \method enable more effective adaptation 
    to novel tasks compared to entangled representations?    
    \item \textbf{Language Grounding Quality (RQ3):} 
    Does \method's decoupled architecture ground 
    the semantic content of instructions -- responding 
    to task meaning rather than surface-level 
    phrasing -- and does the generated parameter 
    space reflect meaningful task structure?    
    \item \textbf{Real-World Deployment (RQ4):} 
    Does \method's decoupling principle transfer 
    to physical robot settings, where real-world perception 
    noise and scene ambiguity might affect precise 
    language-driven control?
\end{enumerate}
The experimental setup 
is detailed in Section~\ref{sec:exp-setup}. 
We report multi-task performance in Section~\ref{sec:multi-task}, few-shot adaptation in Section~\ref{sec:few-shot}, language grounding and structure analysis in Section~\ref{sec:analysis}, and real-world evaluation in Section~\ref{sec:real-world}. 
Further implementation details are provided in the appendix.

\subsection{Experimental Setup}
\label{sec:exp-setup}

\paragraph{Benchmarks}
In simulation, 
we employ two standard benchmarks to test multi-task proficiency: \textbf{LIBERO-90}~\citep{liu2023libero} and \textbf{Meta-World}~\citep{yu2020meta}.
For LIBERO-90, we train on 90 diverse language-conditioned tasks and evaluate under novel initial configurations. To analyze performance relative to task complexity, we categorize these tasks into \textbf{Easy}, \textbf{Medium}, and \textbf{Long-Horizon} groups based on the number of object interactions involved.
For Meta-World, we adopt the ML45 protocol (45 goal-conditioned tasks) and group them by manipulation primitives (e.g., \texttt{pick\&place}, \texttt{press\&pull}) to assess semantic generalization across diverse mechanics.
We utilize the standard dataset budget: 50 human demonstrations per task for LIBERO-90 and 100 expert demonstrations per task for Meta-World.
Detailed task lists and categorization criteria are provided in the appendix.

\begin{figure*}[t]
    \centering
    \includegraphics[width=\linewidth]{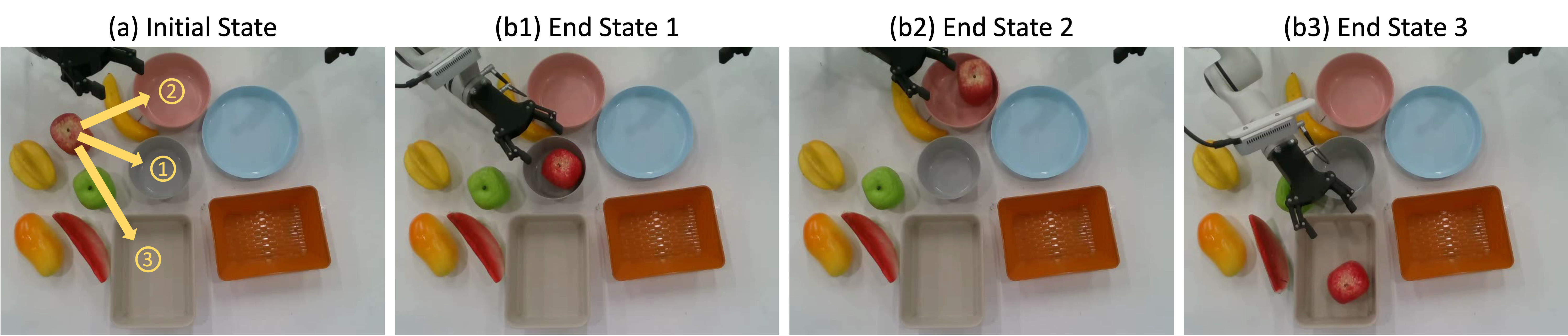} 
    \caption{\textbf{Real-World Combinatorial Benchmark Visualization.} 
    (a) \textbf{Initial State:} The setup presents inherent visual ambiguity: the target object (e.g., Red Apple) has multiple potential destinations (indicated by arrows). 
    (b) \textbf{Language-Conditioned Outcomes:} 
    Starting from the \textit{identical} initial state shown in (a), \method successfully executes three distinct tasks specified by different language instructions: 
    (b1) placing the apple into the \textbf{Small Gray Bowl}, 
    (b2) the \textbf{Large Red Bowl}, and 
    (b3) the \textbf{Light Gray Box}. 
    Because the visual scene is identical in all three cases, the divergent behaviors can only arise from the language instruction -- confirming that the generated policy parameters, not visual context, determine task execution.
    }
    \label{fig:real_world_setup}
    \vspace{-0.3cm}
\end{figure*}

\textbf{Real-World Benchmark.} 
To rigorously evaluate whether language 
instructions -- rather than visual cues -- drive policy behavior, we construct a real-world testbed comprising 9 distinct combinatorial tasks. 
Figure~\ref{fig:real_world_setup} illustrates this experimental protocol.
The setup involves picking one of 3 potential target objects and placing it into one of 3 specific containers, 
resulting in $3 \times 3 = 9$ unique language-conditioned tasks.
As depicted in Figure~\ref{fig:real_world_setup} (a), the workspace is cluttered with distractors. 
The yellow arrows exemplify the task ambiguity that is resolved solely by language: for instance, 
the Red Apple corresponds to three distinct potential trajectories depending on the target container specified in the instruction.
Figure~\ref{fig:real_world_setup} (b) shows the terminal state of a successful execution.
Crucially, because the same objects 
and containers are present in every task, visual context alone cannot determine the correct behavior -- making this benchmark a direct test of whether observation leakage drives the policy. 
We collect 100 demonstrations for each task.
Additional environment details are deferred to the appendix.

\paragraph{Baselines}
We benchmark \method against a suite of state-of-the-art methods, categorized into two groups:
(1) \textbf{Entangled Architectures}, including Transformer-based models such as Octo~\citep{team2024octo} and VQ-BeT~\citep{shafiullah2022behavior}, 
the text-aware VLA model OTTER~\citep{huang2025otter},
and Diffusion-based models like U-Net-based Diffusion Policy (DP)~\citep{chi2023diffusion} and Transformer-based Diffusion (DiT)~\citep{chi2023diffusion}; 
and
(2) \textbf{Hypernetwork-based Architectures}, specifically HyPoGen~\citep{ren2025hypogen} (abbreviated as H-Gen) and HyperZero~\citep{rezaei2023hypernetworks} (H-Zero).
To isolate the impact of architectural inductive biases, 
we train all baselines from scratch using identical visual (ResNet-18) and language (T5-small) backbones.
Architectural details are provided in the appendix.
Furthermore, to contextualize our performance against the current state-of-the-art, 
we include two large-scale pretrained models, $\pi_0$~\citep{black2024pi_0} and $\pi_{0.5}$~\citep{intelligence2025pi_}, fine-tuned on LIBERO-90 as reference points.

\paragraph{Training Protocol}
All experiments follow a standard Behavior Cloning (BC) paradigm.
The aggregate training datasets consist of 4,500 episodes for LIBERO ($50 \times 90$), 4,500 for Meta-World ($100 \times 45$), and 900 for the real-world benchmark ($100 \times 9$).
Crucially, to isolate the impact of proposed architecture from capacity-related benefits, we standardize the model size across all trained-from-scratch methods to $\sim$30M parameters.
For the reference models $\pi_0$ and $\pi_{0.5}$, we fine-tune from their official pretrained checkpoints.
Complete training details are available in the appendix.

\subsection{Multi-Task Performance (RQ1)}
\label{sec:multi-task}

We evaluate whether decoupling instruction from state-conditioned control improves multi-task performance. Models are tested on the same tasks in the training set, but with novel initial configurations (unseen object positions, robot poses) to assess whether the policy generalizes based on language instructions rather than memorized scene-action associations.

\begin{table}[!t]
\centering
\caption{Multi-task performance on LIBERO-90. 
Success rates (\%) averaged over 20 episodes. 
\method shows increasing advantages on complex tasks where 
task-state entanglement is most problematic. 
Best in \textbf{bold}, second-best \underline{underlined} among methods trained from scratch. 
\textit{Italicized values} denote fine-tuned results 
with large-scale pretraining.}
\label{tab:main_libero_transposed}
\setlength{\tabcolsep}{9pt}
\begin{tabular}{lccc|c}
\toprule
Method & Easy & Medium & Long & Overall \\
\midrule
Octo     & 92.3 & \underline{89.0} & 79.3 & 84.7 \\
VQ-BeT   & 89.5 & 86.7 & \underline{84.3} & 85.9 \\
DP    & 93.2 & 72.7 & 72.7 & 75.2 \\
DiT      & 57.3 & 42.0 & 34.2 & 40.1 \\
H-Zero   & 50.0 & 75.1 & 69.1 & 69.1 \\
H-Gen    & 30.5 & 50.7 & 46.3 & 46.1 \\
OTTER    & \underline{96.4} & 87.9 & 83.2 & \underline{86.6} \\
\midrule
\textnormal{DISC} (Ours) & \textbf{97.3} & \textbf{95.3} & \textbf{92.7} & \textbf{94.3} \\
\midrule
\textit{$\pi_0$ (Pretrained + FT)}    & \textit{95.9} & \textit{92.0} & \textit{90.2} & \textit{91.6} \\
\textit{$\pi_{0.5}$ (Pretrained + FT)} & \textit{96.4} & \textit{97.0} & \textit{94.4} & \textit{95.7} \\
\bottomrule
\end{tabular}
\end{table}

\paragraph{Results on LIBERO-90}
Table~\ref{tab:main_libero_transposed} reveals a critical pattern supporting our hypothesis. 
Among trained-from-scratch entangled baselines, OTTER~\citep{huang2025otter} is the strongest overall, reaching 86.6\%, and performs particularly well on Easy tasks with 96.4\%.
\method achieves 94.3\% overall -- a 7.7\% absolute improvement over OTTER. 
When checking the gap against the best from each category, 
we see it correlate with task complexity: 
+0.9\% on Easy tasks (97.3 vs.\ 96.4), 
+6.3\% on Medium tasks (95.3 vs.\ 89.0), 
and +8.4\% on Long-Horizon tasks (92.7 vs.\ 84.3). 
This pattern is consistent with our analysis of task-state entanglement: as task horizon grows, 
the opportunity for observation leakage increases -- 
entangled policies accumulate grounding errors 
across sequential steps because visual shortcuts that work 
for early subgoals may not generalize to later ones. 
\method's decoupled architecture 
avoids this failure mode entirely, 
as the target policy has no mechanism to bypass 
its generated parameters.
Remarkably,
despite being trained entirely
from scratch, \method surpasses the large-scale
pretrained model $\pi_0$ (91.6\%)
and remains competitive with $\pi_{0.5}$ (95.7\%).

\paragraph{Complementary Failure Modes vs.\ $\pi_{0.5}$}
A task-level comparison across all 90 LIBERO tasks reveals that \method and $\pi_{0.5}$ exhibit \emph{complementary} failure modes despite their close aggregate gap: 68 tasks are within 5\%, \method wins by $>$5\% on 9 tasks, and $\pi_{0.5}$ wins by $>$5\% on 13 tasks.
\method's largest advantages appear on \textbf{multi-step tasks requiring precise language-grounded sequencing}: on ``turn on the stove and put the frying pan on it'' across two kitchen scenes, \method achieves 92\%/96\% while $\pi_{0.5}$ drops to 30\%/60\% (\textbf{+62\%}/\textbf{+36\%} gaps), suggesting $\pi_{0.5}$ exploits scene-level visual priors rather than faithfully following the instruction.
This indicates that even internet-scale pretraining does not eliminate observation leakage: action fine-tuning on limited robot data re-introduces shortcut-learning pathways through the entangled fusion of language and visual tokens.
Conversely, $\pi_{0.5}$'s largest advantages appear on \textbf{fine-grained placement tasks} (e.g., ``place book in front compartment''), where \method scores 46--56\% vs.\ $\pi_{0.5}$'s 75--95\%.
Since \method achieves 96--100\% on other book-placement tasks (left/back compartments), the gap reflects motor precision of the compact MLP target policy for specific spatial targets, rather than a language-grounding failure.

\begin{table}[t]
\centering
\caption{Joint evaluation on LIBERO-Spatial, LIBERO-Object, and LIBERO-Goal.
Success rates (\%) are averaged over 50 episodes per task.
$\pi_0$ and $\pi_{0.5}$ are fine-tuned from pretrained checkpoints,
whereas \method is trained from scratch.}
\label{tab:libero_joint_suites}
\setlength{\tabcolsep}{5pt}
\begin{tabular}{lcccc}
\toprule
Method & Spatial & Object & Goal & Overall \\
\midrule
$\pi_0$ & 96.8 & 98.8 & 95.8 & 97.1 \\
$\pi_{0.5}$ & 98.8 & 98.2 & \textbf{98.0} & 98.3 \\
\method & \textbf{99.2} & \textbf{99.4} & 97.4 & \textbf{98.7} \\
\bottomrule
\end{tabular}
\end{table}

\paragraph{Broader LIBERO Suite Evaluation}
We further test whether \method remains competitive on the standard LIBERO-Spatial, LIBERO-Object, and LIBERO-Goal suites under a broader joint-training setting.
As shown in Table~\ref{tab:libero_joint_suites}, \method achieves the highest overall success rate of 98.7\%, slightly exceeding both pretrained reference models despite using no large-scale pretraining.
\method performs best on LIBERO-Spatial (99.2\%) and LIBERO-Object (99.4\%), while $\pi_{0.5}$ obtains the highest score on LIBERO-Goal (98.0\%).
These results provide an additional cross-suite check that the decoupled architecture remains competitive beyond LIBERO-90, rather than being tailored to a single benchmark split.

\begin{table}[t]
\centering
\caption{Meta-World ML45 multi-task performance by manipulation primitives. \method maintains advantages across action types, achieving 92.2\% overall success rate.}
\label{tab:main_mw}
\begin{tabular}{l@{\hspace{0.5em}}c@{\hspace{0.8em}}c@{\hspace{0.8em}}c@{\hspace{0.8em}}c@{\hspace{0.8em}}c@{\hspace{0.8em}}c@{\hspace{0.8em}}c}
\toprule
Task Category & Octo & VQ-BeT & DP & DiT & H-Zero & H-Gen & \method \\
\midrule
open\&close & 99.3 & \textbf{100.0} & 56.8 & 52.7 & 99.3 & 91.5 & \underline{99.5} \\
pick\&place & \underline{83.0} & 69.0 & 19.5 & 19.0 & 81.5 & 66.5 & \textbf{88.0} \\
press\&pull & \textbf{93.4} & \underline{88.1} & 33.9 & 38.5 & 38.4 & 72.5 & \textbf{93.4} \\
others & \underline{85.8} & 77.6 & 54.6 & 48.1 & 48.2 & 68.5 & \textbf{88.7} \\
\midrule
Overall & \underline{90.6} & 84.6 & 44.5 & 42.9 & 56.8 & 73.8 & \textbf{92.2} \\
\bottomrule
\end{tabular}
\end{table}

\paragraph{Results on Meta-World} 
Meta-World results (Table~\ref{tab:main_mw}) confirm that \method's advantages generalize across benchmarks. 
While simple \texttt{open\&close} tasks saturate near 100\% for most methods, \method shows clear improvements on tasks requiring precise object manipulation (\texttt{pick\&place}: +5.0\% over Octo) and complex action sequences 
(\texttt{others}: +2.9\%). 
The consistent performance across categories demonstrates that decoupling instruction from state-conditioned control benefits diverse manipulation primitives across task categories, not just specific tasks in isolation.

\subsection{Few-Shot Adaptation to Unseen Tasks (RQ2)}
\label{sec:few-shot}

\begin{figure}[t]
    \centering
    \includegraphics[width=1.0\linewidth]{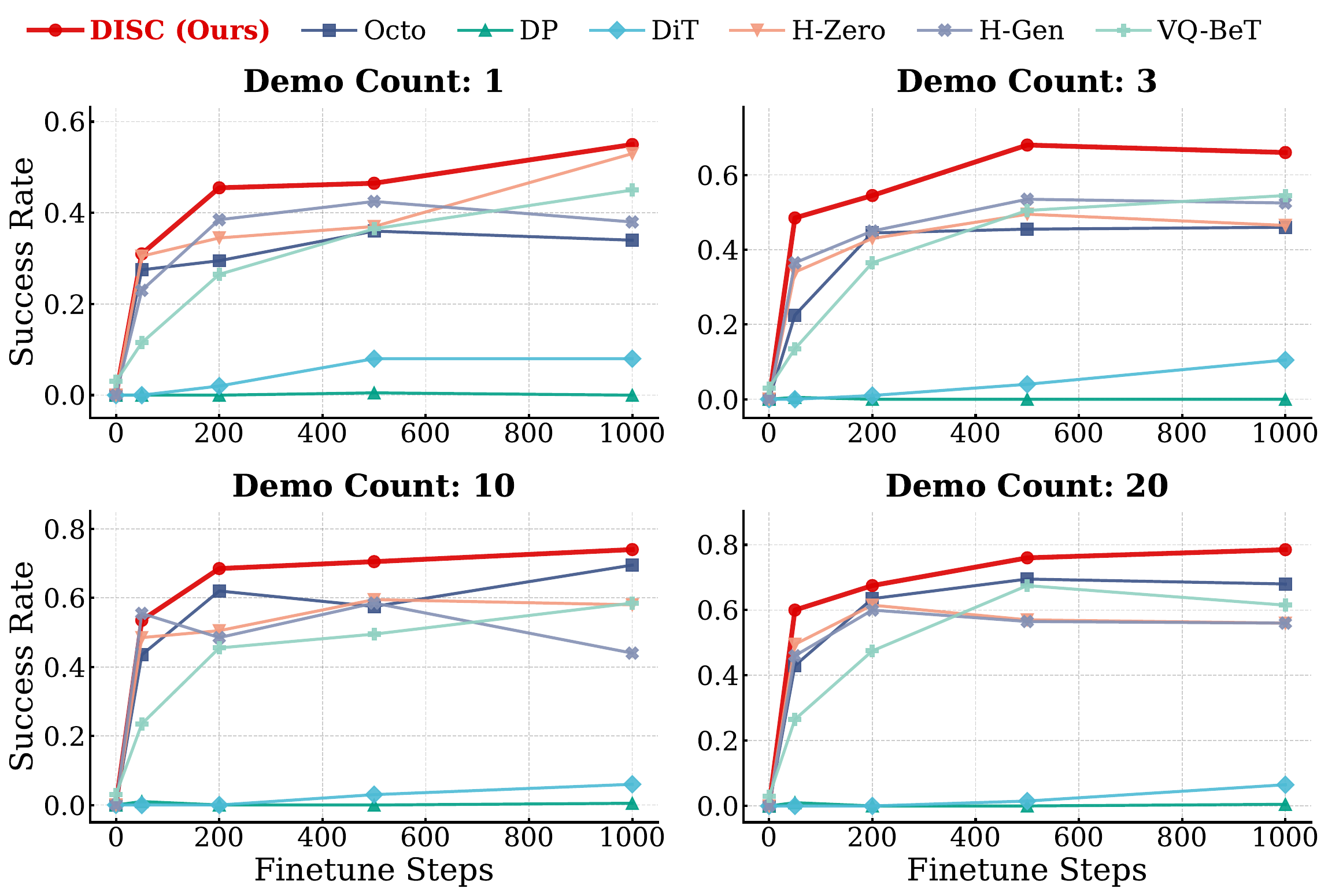}
    \caption{Few-Shot Adaptation Efficiency on LIBERO-Spatial. 
    The plots show success rate over 1000 fine-tuning steps, with $K \in \{1, 3, 10, 20\}$ demonstrations. 
    \method shows better sample efficiency than state-of-the-art baselines like Octo and DP.}
    \label{fig:tta_spatial_results}
    \vspace{-0.3cm}
\end{figure}

We evaluate the adaptation efficiency of \method on held-out LIBERO-Spatial tasks. 
This experiment assesses whether the structured parameter manifold learned by our hypernetwork enables efficient transfer to novel tasks under varying data budgets ($K \in \{1, 3, 10, 20\}$ demonstrations).

\paragraph{Adaptation Protocol}
Our adaptation protocol ensures a rigorous comparison across architectures.
For Hypernetwork-based methods, we query the frozen hypernetwork with the new instruction to generate a policy initialization $\theta_{\pi}^{\text{init}}$, then fine-tune only the generated parameters.
For entangled baselines, we adopt the standard LoRA~\citep{hu2022lora} strategy. 
To maintain fairness, the trainable parameter count for LoRA adapters is matched ($\approx$ 0.74M) to that of our generated policies.
All methods are optimized using the same loss and optimizer for 1,000 gradient steps. 
We report the success rate evaluated at key checkpoints across different fine-tuning steps, averaged over 20 evaluation episodes per task.

\paragraph{Results and Analysis}
Figure~\ref{fig:tta_spatial_results} highlights \method's superior sample efficiency across diverse data regimes. 
Notably, in the \textbf{extreme few-shot setting ($K=1, 3$)}, \method achieves non-trivial success rates where diffusion-based methods (DP, DiT) struggle near 0\% performance. With very few demonstrations, entangled architectures lack sufficient data to disentangle task specification from state observation, causing the optimization to collapse onto spurious visual correlations. \method's decoupled architecture sidesteps this failure: because the hypernetwork has already encoded task semantics into the generated initialization, fine-tuning only needs to adjust visuomotor precision rather than learn task identity from scratch.
Furthermore, \method demonstrates \textbf{rapid convergence}: as seen in the $K=10$ and $K=20$ plots, our performance curve rises sharply within the first 200 steps, significantly outpacing baselines like Octo which exhibit a slower ``warm-up'' phase. 
This confirms that the hypernetwork provides a semantically informed initialization near the task-optimal region, transforming adaptation from a global search over an entangled parameter space into a local refinement of a compact, well-initialized policy. See the appendix for additional results.

\subsection{Language Grounding and Structure Analysis (RQ3)}
\label{sec:analysis}

\paragraph{Semantic Grounding under Linguistic Variation}
A central claim of \method is that the hypernetwork learns to extract task semantics from language, not memorize surface-level token patterns. To test this, all methods are retrained under identical conditions using augmented language data: 50 paraphrased instructions per task (e.g., ``pick up the red block'' $\rightarrow$ ``grab the crimson cube''), and then evaluated on 10 additional held-out descriptions not seen during training. This measures each architecture's ability to map diverse phrasings of the same task to functionally equivalent control behavior -- a test of whether the model grounds task meaning rather than memorizes specific wording.
As shown in Table~\ref{tab:language-robust}, \method achieves the highest performance on both benchmarks: 85.4\% on LIBERO-90 (vs.\ Octo's 80.6\%) and 91.8\% on Meta-World (vs.\ Octo's 90.3\%).
This result follows naturally from \method's architecture. Because the hypernetwork must decode an instruction into a \emph{complete set of target policy parameters} -- the only channel through which language influences the task-specific action mapping -- it is incentivized to extract the underlying task specification rather than relying on specific lexical patterns. 
In entangled architectures, language tokens and visual tokens share processing stages, allowing the network to form spurious associations between particular word tokens and visual features. When phrasing changes, these associations break.

\begin{table}[t]
\centering
\caption{Semantic grounding under linguistic variation. All methods are retrained with paraphrased data and evaluated on held-out descriptions. \method maintains higher success rates, indicating that the decoupled architecture grounds task meaning rather than memorizes specific phrasings.}
\label{tab:language-robust}
\setlength{\tabcolsep}{5pt}
\begin{tabular}{lccccccc}
\toprule
Dataset & Octo & VQ-BeT & DP & DiT & H-Zero & H-Gen & \method \\
\midrule
LIBERO-90 & \underline{80.6} & 76.4 & 34.5 & 29.7 & 66.4 & 45.9 & \textbf{85.4} \\ 
Meta-World & \underline{90.3} & 85.9 & 49.5 & 23.3 & 79.2 & 83.1 & \textbf{91.8} \\
\bottomrule
\end{tabular}
\end{table}

\begin{figure}
\centering
\includegraphics[width=\linewidth]{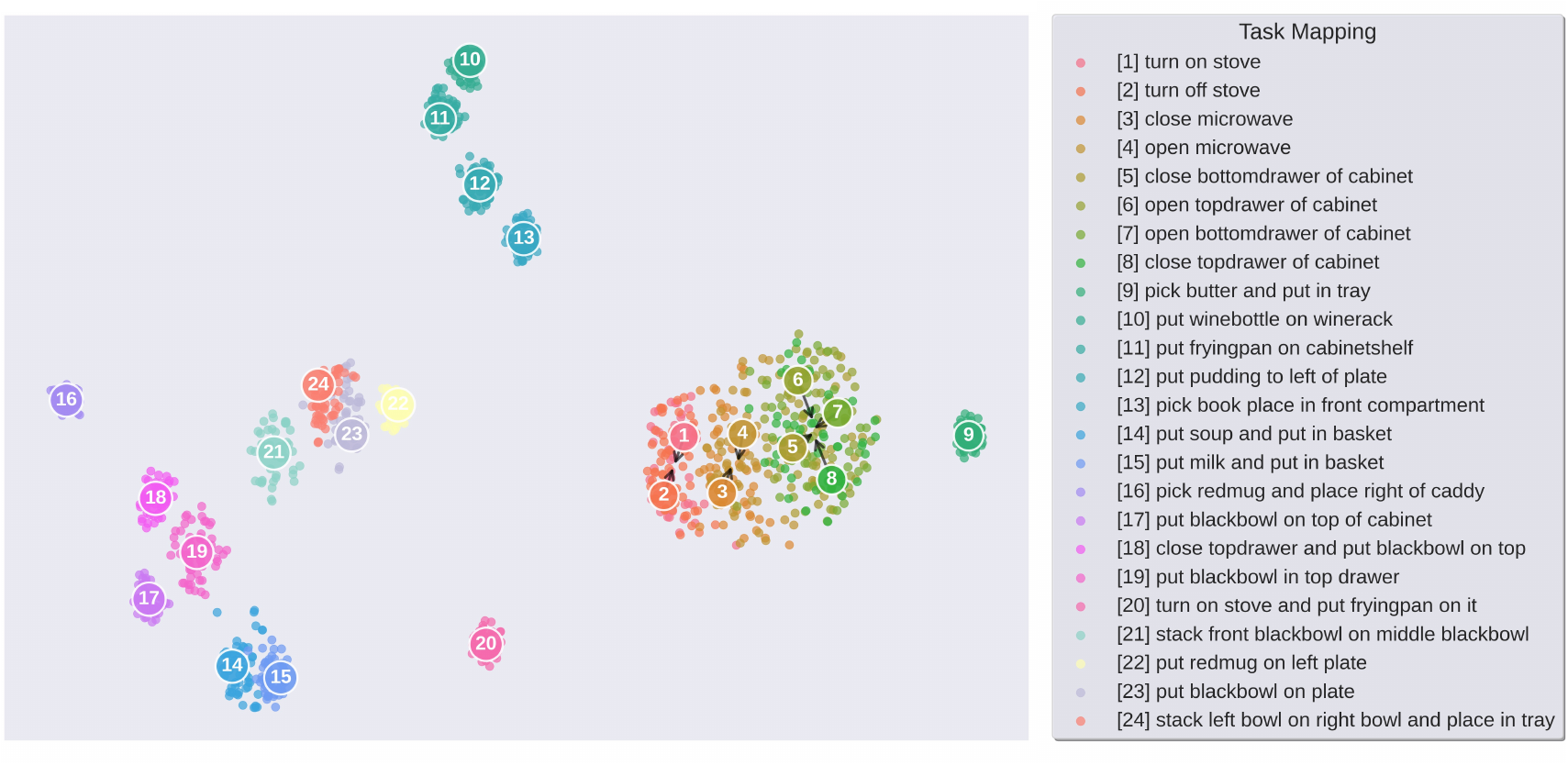}
\caption{t-SNE visualization of generated policy parameters. 
Semantically similar tasks cluster based on functional meaning, 
indicating that the hypernetwork maps linguistic task specifications 
to a structured parameter manifold.}
\label{fig:tsne}
\vspace{-0.3cm}
\end{figure}

\paragraph{Parameter Space Visualization}
Figure \ref{fig:tsne} visualizes the generated parameter manifold via t-SNE~\citep{maaten2008visualizing}, revealing that the hypernetwork organizes policy parameters by functional semantics rather than visual similarity.
We observe three key properties:
(1) \textbf{Action-Type Separation:} A macroscopic distinction exists between \textit{articulated object manipulation} (Tasks 1--8, central cluster) and \textit{pick-and-place} tasks (Tasks 10--24, periphery), indicating the capture of high-level control primitives in the generated weights.
(2) \textbf{Functional Grouping:} The central cluster further separates tasks by object affordance, distinguishing Cabinet/Drawer interactions (Tasks 5--8) from Stove and Microwave operations.
(3) \textbf{Semantic Continuity:} Inverse operations, such as Turn On/Off Stove (Tasks 1--2) and Open/Close Microwave (Tasks 3--4), are mapped as nearest neighbors.
This topology confirms that the hypernetwork has learned a mapping where semantic proximity in language corresponds to geometric proximity in the weight manifold. This structure directly explains the few-shot adaptation advantages observed in Section~\ref{sec:few-shot}: novel tasks with semantically similar instructions receive nearby initializations, reducing the distance that fine-tuning must traverse.

\subsection{Real-World Deployment (RQ4)}
\label{sec:real-world}

The simulation experiments establish \method’s advantages under controlled conditions. We now evaluate whether the decoupling principle transfers to physical robot settings, where richer visual correlations, increased perceptual noise, and narrower grounding margins exacerbate observation leakage.

\begin{table}[t]
\centering
\caption{Real-world performance on 9 combinatorial tasks. The setup involves picking one of 3 target objects into one of 3 containers in a cluttered scene where visual context is shared across all tasks. Success rates (\%) averaged over 30 episodes. (Abbr: R.A.=Red Apple, G.A.=Green Apple, W.M.=Watermelon; S.G.=Small Gray, L.R.=Large Red, L.G.=Light Gray).}
\label{tab:real_world}
\setlength{\tabcolsep}{3.5pt}
\resizebox{\linewidth}{!}{
\begin{tabular}{lccccc}
\toprule
ID & Instruction (Obj $\to$ Container) & H-Zero & DP & Octo & \textbf{\method} \\
\midrule
01 & R.A. $\to$ S.G. Bowl & 73.3 & 60.0 & 73.3 & \textbf{96.7} \\
02 & R.A. $\to$ L.R. Bowl & 10.0 & 23.3 & 53.3 & \textbf{83.3} \\
03 & R.A. $\to$ L.G. Box & 13.3 & 33.3 & 70.0 & \textbf{90.0} \\
04 & G.A. $\to$ S.G. Bowl & 3.3 & 43.3 & 80.0 & \textbf{93.3} \\
05 & G.A. $\to$ L.R. Bowl & 6.7 & 56.7 & \textbf{73.3} & 53.3 \\
06 & G.A. $\to$ L.G. Box & 0.0 & 76.7 & 83.3 & \textbf{100.0} \\
07 & W.M. $\to$ S.G. Bowl & 36.7 & 53.3 & \textbf{90.0} & \textbf{90.0} \\
08 & W.M. $\to$ L.R. Bowl & 16.7 & \textbf{96.7} & 93.3 & 86.7 \\
09 & W.M. $\to$ L.G. Box & 26.7 & 66.7 & \textbf{96.7} & 83.3 \\
\midrule
\multicolumn{2}{c}{\textbf{Average Success Rate (\%)}} & 18.5 & 57.0 & 78.5 & \textbf{86.4} \\
\bottomrule
\end{tabular}}
\end{table}

We evaluate \method on the combinatorial benchmark described in Section~\ref{sec:exp-setup}, which provides a direct test of observation leakage on a physical robot.
The 9 tasks are formed by the Cartesian product of 3 distinct objects (Red Apple, Green Apple, Watermelon) and 3 target containers (Small Gray Bowl, Large Red Bowl, Light Gray Box), all presented simultaneously in a cluttered workspace with distractors.
This setting is particularly revealing: because the same objects and containers are present in every task, visual context is nearly identical across all 9 tasks. A policy that has learned scene-to-action mappings through observation leakage will systematically confuse tasks that share visual context, while a policy whose behavior is determined by language-generated policy parameters will distinguish them correctly.

As shown in Table~\ref{tab:real_world}, \method achieves the highest average success rate, outperforming all baselines by clear margins.
Qualitative analysis reveals distinct failure modes that align with our theoretical framework.
HyperZero frequently fails due to \textit{spatial precision errors} (e.g., ``air grasps''), indicating that without the iterative refinement stage, the generated parameters lack sufficient fidelity for precise physical control -- validating the architectural contribution of our optimization-inspired refinement module.
Entangled baselines exhibit a qualitatively different failure: Octo and DP often achieve correct localization but suffer from \textit{orientation misalignment}, suggesting that their shared representations have learned coarse scene-to-action associations rather than precise language-driven control.
This is the signature of observation leakage: the policy partially succeeds by mapping visual context to approximate actions, but fails when the task requires fine-grained distinctions that only language can resolve.
Notably, Octo performs well on visually distinctive tasks (e.g., Task~09, Watermelon $\to$ Light Gray Box) but degrades sharply on tasks requiring subtle container distinctions (e.g., Task~02, Red Apple $\to$ Large Red Bowl vs.\ Small Gray Bowl) -- precisely the failure mode predicted when visual context rather than language determines behavior.
\method mitigates both failure modes:
the decoupled architecture ensures that task identity is encoded entirely in the generated parameters, eliminating observation leakage, while the refinement module provides the parameter precision needed for reliable physical execution.

\paragraph{Real-World Failure Analysis}
\method's only sub-baseline result is on Task~05 (G.A.\ $\to$ L.R.\ Bowl, 53.3\%), where the gripper tends to grasp the apple at suboptimal positions, leading to drops during transport.
Notably, \method achieves 93.3\% and 100\% on the other two Green Apple tasks (04, 06) and 83.3\% on the same target container with a different object (Task~02), confirming that the failure stems from execution-level imprecision rather than a language-grounding error.
This indicates that the current bottleneck is the compact MLP target policy, which can be insufficient for physically demanding placements.
Generating more expressive target policies (e.g., diffusion- or flow-based) is a promising direction for further improving precision and robustness.

\noindent\textbf{Summary.} 
Across simulation and real-world settings, our experiments confirm that decoupling instruction from state-conditioned control yields consistent improvements in multi-task performance, few-shot adaptation, semantic grounding, and physical deployment. These advantages scale with task complexity and visual ambiguity -- precisely the conditions under which task-state entanglement is most damaging, and where \method's architectural guarantees are most valuable.

\section{Conclusion}

The central finding of this work 
is that \emph{where} language enters a policy architecture 
matters more than \emph{how} it is encoded. 
Task-state entanglement is 
not merely a training problem solvable with more data or larger models 
-- it is a structural property 
of shared-parameter designs. 
\method demonstrates that generating task-specific policy parameters from language blocks the shared action-mapping pathway behind this entanglement, recovering instruction grounding, generalization, and few-shot adaptation beyond what scaling entangled architectures has achieved. 
The learned parameter manifold organizes tasks as geometric neighbors in weight space without explicit supervision, 
suggesting that decoupling unlocks structure entangled architectures cannot express. 
In practice, \method is most attractive when reliable grounding, efficient control, and adaptation from limited robot data are primary requirements. 
Large pretrained VLAs remain complementary: they are well suited to data-rich and compute-rich settings where broad pretraining can be fully exploited. 

\noindent\textbf{Limitations.}
The main practical trade-off of \method lies on the training side, where it can require more memory and longer optimization than comparable static policies.
Under-specified instructions may also surface uncertainty rather than being silently resolved through observation leakage, motivating future work on more efficient refinement, uncertainty-aware clarification, and broader multi-modal extensions.

\section*{Acknowledgments}
This work is supported by 
the Early Career Scheme 
of the Research Grants Council (RGC) grant \# 27207224,
the HKU-100 Award, 
a donation from the Musketeers Foundation, 
and in part by the 
JC STEM Lab of 
Autonomous Intelligent Systems 
funded by 
The Hong Kong Jockey Club Charities Trust.


\bibliographystyle{plainnat}
\bibliography{references}

\clearpage
\appendices

\section{Problem Formulation Details}
\label{app:problem_formulation}

\subsection{State and Action Spaces}
The state space $\mathcal{S} \subseteq \mathbb{R}^{H \times W \times C} \times \mathbb{R}^{d_s}$ consists of:
\begin{itemize}
    \item RGB images $I_t \in \mathbb{R}^{H \times W \times C}$ with resolution $H \times W$ and $C=3$ color channels. Specifically, in LIBERO-90 experiments, an additional gripper-view image is also included.
    \item Proprioceptive state $s_t \in \mathbb{R}^{d_s}$ including end-effector pose and gripper state.
\end{itemize}

The continuous action space $\mathcal{A} \subseteq \mathbb{R}^{d_a}$ represents target joint velocities or end-effector displacements, depending on the control mode. The transition dynamics $\mathcal{T}: \mathcal{S} \times \mathcal{A} \to \Delta(\mathcal{S})$, where $\Delta(\mathcal{S})$ denotes the probability simplex over $\mathcal{S}$, capture the stochastic nature of manipulation tasks.

\subsection{Dataset Structure}
Each demonstration trajectory $\xi_j^i$ has a horizon $T_j^i$ and consists of:
\begin{itemize}
    \item Observations $o_t^{i,j} = (I_t^{i,j}, s_t^{i,j})$ combining visual and proprioceptive information.
    \item Expert actions $a_t^{i,j} \in \mathcal{A}$ recorded from human demonstrations or privileged controllers.
    \item Language instruction $l_i$ describing the task goal in natural language.
\end{itemize}

\section{Examples of Task-State Entangled Architectures}
\label{app:task-state-entangled}

Modern language-conditioned robotic policies employ various forms of task-state entanglement:

\textbf{Transformer-based Fusion.} Policies like Octo~\citep{team2024octo} tokenize both language and visual inputs, concatenating them into a single sequence for processing. The self-attention mechanism allows arbitrary interactions between language and visual tokens, creating entangled representations throughout the network depth.

\textbf{Diffusion-based Fusion.} Diffusion Policies~\citep{chi2023diffusion} typically incorporate language by concatenating embeddings with visual features or inputs, followed by architecture-specific conditioning mechanisms:
\begin{itemize}
    \item \textbf{FiLM conditioning:} Language embeddings modulate visual features through affine transformations when using U-Net-based diffusion policies.
    \item \textbf{Cross-attention:} Action denoising queries attend to both language and visual keys when using transformer-based diffusion policies (DiT).
\end{itemize}

In all cases, the action prediction emerges from representations where language and vision have been deeply intertwined through non-linear transformations, making it difficult to isolate their individual contributions.

\section{Environmental Setup and Baselines}
\label{app:experimental_details}

\subsection{Task Statistics Overview}
We evaluate on two benchmarks: LIBERO-90 and Meta-World ML45.

\noindent\textbf{LIBERO-90:} Contains 90 tasks in total.
\begin{itemize}
    \item \textbf{Easy (11 tasks):} Single-object manipulation.
    \item \textbf{Medium (35 tasks):} Multi-object coordination.
    \item \textbf{Long (44 tasks):} Sequential multi-step goals.
\end{itemize}
Please refer to Appendix~\ref{app:libero-splits} for the complete list of all 90 tasks.

\noindent\textbf{Meta-World ML45:} Contains 45 training tasks covering pick-place, push, press, and open-close interactions. The full list is provided in Appendix~\ref{app:metaworld-splits}.

\subsection{Sampling Strategy}
For all the experiments, we first construct a global list of all transitions across the dataset, where each entry is a tuple $(\text{task}, \text{episode}, \text{timestep})$. At training time, each batch is formed by uniformly sampling transitions from this list.

For language augmentation experiments in the main paper, we use the same transition-level sampling strategy. In addition, for each sampled transition we randomly draw a paraphrased instruction from that task’s pool of 50 paraphrases, so the model sees diverse linguistic realizations of the same underlying task during training.

\subsection{Baseline Implementation Details}
We compare against six state-of-the-art methods across two categories:

\noindent\textbf{Task-state Entangled Methods:}
\begin{itemize}[leftmargin=*, topsep=2pt, itemsep=1pt]
    \item \textbf{Octo}~\citep{team2024octo}: Transformer processing concatenated language-vision tokens with self-attention mechanism.
    \item \textbf{VQ-BeT}~\citep{shafiullah2022behavior}: Vector-quantized behavior transformer using self-attention mechanism.
    \item \textbf{Diffusion Policy}~\citep{chi2023diffusion}: UNet-based diffusion with FiLM conditioning for language and observation information.
    \item \textbf{DiT}~\citep{chi2023diffusion}: Diffusion transformer using cross-attention for language and observation information.
\end{itemize}

\noindent\textbf{Hypernetwork Methods:}
\begin{itemize}[leftmargin=*, topsep=2pt, itemsep=1pt]
    \item \textbf{HyPoGen}~\citep{ren2025hypogen}: Iterative hypernetwork with optimization bias to decode the policy parameters from the language embedding.
    \item \textbf{HyperZero}~\citep{rezaei2023hypernetworks}: Direct MLP mapping from language to parameters to decode the policy parameters from the language embedding.
\end{itemize}

All baselines are implemented in PyTorch with the following specifications:

\noindent\textbf{Shared Components:}
\begin{itemize}[topsep=2pt, itemsep=1pt]
    \item Visual encoder: ResNet-18 pretrained on ImageNet, fine-tuned during training.
    \item Language encoder: T5-small encoder~\citep{raffel2020exploringt5}, and is kept frozen during training in LIBERO-90 and Meta-World experiments.
    \item Action space: 6-DoF end-effector delta and 1-dim state of gripper for LIBERO, 3-DOF end-effector delta and 1-dim gripper for Meta-World.
\end{itemize}

\noindent\textbf{Model-Specific Details:}
\begin{itemize}[topsep=2pt, itemsep=1pt]
    \item \textbf{Octo:} 8-layer transformer, 8 attention heads, hidden dim 512.
    \item \textbf{VQ-BeT:} Codebook size 16 with 2 groups, 8-layer transformer, 8 attention heads, hidden dim 488.
    \item \textbf{Diffusion Policy:} 100 denoising steps, DDIM sampler, FiLM conditioning, down-scale dims [128, 256, 512], and timestep embedding dim 256.
    \item \textbf{DiT:} 8-layer diffusion transformer, cross-attention for language and observation.
    \item \textbf{HyPoGen:} 3 refinement blocks, hidden dim 32.
    \item \textbf{HyperZero:} 5-layer MLP hypernetwork, hidden dim 48, ReLU activations.
\end{itemize}

For all hypernet-based methods, we use the same policy network of a 5-layer MLP with hidden dim 320, resulting in a total parameter of around 0.74M in LIBERO and real-world experiments, and 0.57M in Meta-World experiments.

To ensure a fair comparison, we maintain a consistent model capacity across all methods above, with approximately 30M trainable parameters.

\noindent\textbf{Large-scale Pretrained Reference Models:}
\begin{itemize}[topsep=2pt, itemsep=1pt]
    \item \textbf{$\pi_0$}: The official Pi-0 checkpoint~\citep{black2024pi_0}, fine-tuned on LIBERO-90 to serve as a high-performance baseline.
    \item \textbf{$\pi_{0.5}$}: The official Pi-0.5 checkpoint~\citep{intelligence2025pi_}, fine-tuned on LIBERO-90 to serve as an additional reference point.
\end{itemize}

\section{Training and Implementation Details}
\label{app:implementation}

\subsection{Language Encoder}
We employ T5-small~\citep{raffel2020exploringt5} as our language encoder $\Phi_L$, mapping natural language instructions into 512-dimensional embeddings. Instructions are capped at a maximum length of 32 tokens. To bridge the dimension gap, an additional learned projection network aligns these language embeddings with the hypernetwork's task-conditioning space.

\subsection{Neural Gradient Estimation and Update Module}
The iterative refinement module (Stage 2) employs a sophisticated neural gradient estimation mechanism inspired by meta-learning. It is important to note that this is not a true optimization process, but a learned process that mimics the optimization process. At each refinement step $t \in \{0, \dots, T-1\}$, the module simulates an optimization process through three sub-networks:

\textbf{Forward Pass Simulation:} The module first simulates a forward pass through the target network to estimate task-specific activation patterns:
\begin{equation}
(z_0, \dots, z_n) = F_{\text{Forward}}(\theta_\pi^t, e_l; \phi_{2,f})
\end{equation}
where $z_i$ represents estimated activations at layer $i$ and $\phi_{2,f}$ are the forward simulator parameters.

\textbf{Backward Pass Simulation:} Next, it simulates a backward pass to compute pseudo-gradients:
\begin{equation}
\left(\frac{\partial L}{\partial z_n}, \dots, \frac{\partial L}{\partial z_1}\right) = F_{\text{Backward}}(\theta_\pi^t, e_l, z_0, \dots, z_n; \phi_{2,b})
\end{equation}
These pseudo-gradients are not true gradients of an explicit loss function but learned signals that guide parameter updates toward task-specific optima.

\textbf{Meta-Update Network:} Finally, a meta-update network computes the parameter update:
\begin{equation}
\Delta\theta^t = F_{\text{Meta}}\left(\theta_\pi^t, e_l, \frac{\partial L}{\partial z_n}, \dots, \frac{\partial L}{\partial z_1}; \phi_{2,m}\right)
\end{equation}
The complete refinement module parameters are denoted as $\phi_2 = \{\phi_{2,f}, \phi_{2,b}, \phi_{2,m}\}$. This design allows the hypernetwork to learn task-specific optimization trajectories in parameter space, effectively performing learned optimization tailored to each language instruction.

\textbf{Attention-Based Implementation Details:} Our forward and backward pass simulators leverage cross-attention mechanisms to model neural gradient computation through tokenized representations. We tokenize parameter matrices $\theta_i$ of layer $i$ row-wise as $\boldsymbol{\omega}_i = \{\omega_1, \omega_2, \dots, \omega_{n_i}\}$ using MLP encoders, while activations $z_i$ are tokenized as $\boldsymbol{\tau}_i = \{\tau_1, \tau_2, \dots, \tau_{n_i}\}$ representing individual neurons. We use a uniform $d=128$ for all token representations in our model. The forward simulator $F_{\text{Forward}}$ implements layer-wise computation as $\boldsymbol{\tau}_i = \text{CrossAttn}(\boldsymbol{\omega}_{i}, \boldsymbol{\tau}_{i-1}, \boldsymbol{\tau}_{i-1})$, where parameter tokens serve as queries attending over previous layer activations with attention head dimension $d_k = 128$. For backward pass simulation, $F_{\text{Backward}}$ estimates Jacobians using $J_{\theta_i} = \text{CrossAttn}(\boldsymbol{\tau}_{i-1}, \boldsymbol{\omega}_{i}, \boldsymbol{\omega}_{i})$ and $J_{h_i} = \text{CrossAttn}(\boldsymbol{\omega}_{i}, \boldsymbol{\tau}_{i-1}, \boldsymbol{\tau}_{i-1})$ for inter-layer dependencies. Chain rule computations are implemented via attention-based matrix multiplications: $\partial L / \partial z_{i-1} = \text{CrossAttn}(\partial L / \partial z_{i}, J_{h_i}, J_{h_i})$, ensuring modality consistency by using upstream gradients as queries while Jacobian estimates serve as both keys and values. Each cross-attention module employs 4 attention heads with 1 layer, enabling the model to capture fine-grained dependency structures essential for effective parameter generation.

\subsection{Architectural Design Choices}
The WIN can be implemented as either a multi-layer perceptron (MLP) or a Transformer, depending on the complexity of the target network. In our experiments, we employ a 4-layer transformer with cross-attention blocks for the WIN.

\begin{algorithm}
\caption{Single Step Update for \method}
\label{alg:single-step-update}
\begin{algorithmic}[1]
\State \textbf{Input:} Parameters $\theta^{(t-1)}$, task embedding $e_l$
\State \textbf{Output:} Updated parameters $\Delta\theta^{(t-1)}$
\State \textbf{Tokenize parameters for each layer $i$'s parameter $\theta_i$:}
\State \quad $\omega_i \gets \text{Tokenize}(\theta_i)$
\State \textbf{Compute initial activation $\tau_0$:}
\State \quad $\tau_0 \gets \text{CrossAttn}(\text{learnable query}, e_l, e_l)$
\State \textbf{Forward Pass Computation Using Cross-Attention:}
\State \quad $\tau_i \gets \text{CrossAttn}(\omega_i, \tau_{i-1}, \tau_{i-1})$
\State \textbf{Backward Pass and Jacobian Estimation:}
\State \quad $\frac{\partial z_i}{\partial z_{i-1}} \gets \text{CrossAttn}(\omega_i, \tau_{i-1}, \tau_{i-1})$
\State \quad $\frac{\partial z_i}{\partial \theta_i} \gets \text{CrossAttn}(\tau_{i-1}, \omega_i, \omega_i)$
\State \textbf{Gradient Computation:}
\State \quad $\frac{\partial L}{\partial z_{i-1}} \gets \text{CrossAttn}\left(\frac{\partial L}{\partial z_i}, \frac{\partial z_i}{\partial z_{i-1}}, \frac{\partial z_i}{\partial z_{i-1}}\right)$
\State \quad $\nabla \omega_i \gets \text{CrossAttn}\left(\frac{\partial L}{\partial z_i}, \frac{\partial z_i}{\partial \theta_i}, \frac{\partial z_i}{\partial \theta_i}\right)$
\State \textbf{Decode into $\theta$ space}
\State \quad $\Delta \theta_i \gets \text{MLP}(\nabla \omega_i) \text{ and proper reshape}$
\end{algorithmic}
\end{algorithm}

\subsection{Training Details}
\noindent\textbf{Visual Data Augmentation:}
During training, we apply standard visual augmentations including random resized crops and color jittering to the input RGB images to improve policy robustness.

\noindent\textbf{Optimization:}
\begin{itemize}[topsep=2pt, itemsep=1pt]
    \item \textbf{Optimizer:} We use the AdamW optimizer with a learning rate of $1\times 10^{-4}$, $\beta_1=0.9$, $\beta_2=0.999$, and a weight decay of $1\times 10^{-4}$.
    \item \textbf{Learning Rate Schedule:} A cosine annealing schedule is employed throughout the training.
    \item \textbf{Training Stability:} For DiT, we apply gradient clipping with a max norm of 1.0 to ensure stability. Other models do not require gradient clipping.
    \item \textbf{Precision:} We utilize mixed-precision training (BF16) for all models to accelerate computation, with the exception of DiT, which is trained in FP32 to avoid numerical instability.
\end{itemize}

\section{Training \& Inference Cost Analysis}
\method follows a \textbf{``Generate Once, Act Many''} paradigm: the hypernetwork generates the policy weights once per task/language instruction, after which only the lightweight Target Network is used in the control loop. In Table \ref{label:compute-cost}, we show the computation cost comparison between \method with other baselines. \method and HyPoGen share the same lightweight target network, so their control-loop inference speed is effectively identical, both running comfortably in the sub-millisecond regime. Both hypernetwork-based methods are substantially faster than monolithic architectures like Octo, and orders of magnitude faster than diffusion-based policies such as Diffusion Policy. Although \method employs a more expressive hypernetwork and its weight generation is correspondingly slower, this cost remains real-time and is incurred only once per task. As a result, the weight-generation overhead is negligible in the overall control budget, while \method still enjoys the efficiency benefits of an extremely fast controller.

\begin{table}[ht]
\small
\centering
\caption{Computational cost breakdown for \method and baselines on an NVIDIA RTX 4090 GPU.}
\label{label:compute-cost}
\begin{tabularx}{0.48\textwidth}{Xcc}
\toprule
\textbf{Component / Model} & \textbf{Time (ms)} & \textbf{FPS} \\
\midrule
\method Weight Gen. \\
\hspace{1em}\textit{\footnotesize (Full hypernet run)} & 15.75 & $\sim 64$ \\
\addlinespace[0.3em]
\method Target Net \\
\hspace{1em}\textit{\footnotesize (Policy execution)} & 0.13 & $\sim 7422$ \\
\addlinespace[0.3em]
HyPoGen Weight Gen. \\
\hspace{1em}\textit{\footnotesize (Full hypernet run)} & 3.38 & $\sim 295$ \\
\addlinespace
\hline
\multicolumn{3}{l}{\textbf{End-to-End Control Loop}} \\
\addlinespace[0.2em]
\method (Ours) \\
\hspace{1em}\textit{\footnotesize (Fastest control loop)} & 0.67 & $\sim 1485$ \\
\addlinespace[0.3em]
HyPoGen \\
\hspace{1em}\textit{\footnotesize (Same target net)} & 0.67 & $\sim 1485$ \\
\addlinespace[0.3em]
Octo \\
\hspace{1em}\textit{\footnotesize (One-pass transformer)} & 1.86 & $\sim 539$ \\
\addlinespace[0.3em]
Diffusion Policy \\
\hspace{1em}\textit{\footnotesize (Multi-step denoising)} & 160.17 & $\sim 6$ \\
\bottomrule
\end{tabularx}
\end{table}

\method and HyPoGen share the same lightweight target network, so their control-loop inference speed is effectively identical. Both are substantially faster than monolithic architectures like Octo, and orders of magnitude faster than diffusion-based policies.

\begin{table*}[h]
  \centering
  \caption{Success rates (\%) on the LIBERO-10 suite. Bold indicates the best result; underline indicates the second best, and the same convention applies hereafter.}
  \label{tab:libero_10_transposed}
  \begin{tabular}{lcccccccc}
    \toprule
    \textbf{{Demos}} & \textbf{{Steps}} & Octo & VQ-BeT & DP & DiT & \underline{H-Zero} & H-Gen & \textbf{\method} \\
    \midrule
    \textbf{1} & 0 & 3.5 & 3.0 & 3.0 & 1.0 & 0.0 & 0.0 & 3.0 \\
    & 50 & 12.5 & 9.0 & 1.0 & 2.0 & 13.5 & 3.5 & 16.0 \\
    & 200 & 9.5 & 15.5 & 0.0 & 0.0 & 10.0 & 14.0 & 16.0 \\
    & 500 & 10.5 & 11.0 & 0.0 & 1.0 & 13.0 & 19.0 & 24.0 \\
    & 1000 & 14.0 & 10.0 & 2.0 & 2.0 & 29.0 & 22.0 & 20.5 \\
    \midrule
    \textbf{3} 
    & 50 & 17.0 & 8.5 & 2.0 & 2.0 & 16.5 & 16.0 & 13.5 \\
    & 200 & 24.5 & 13.5 & 0.0 & 1.5 & 22.0 & 27.0 & 22.5 \\
    & 500 & 16.0 & 19.5 & 0.0 & 5.5 & 28.5 & 27.0 & 30.0 \\
    & 1000 & 17.5 & 25.5 & 0.0 & 3.0 & 34.5 & 26.0 & 30.0 \\
    \midrule
    \textbf{5} 
    & 50 & 11.5 & 11.0 & 2.0 & 1.0 & 23.0 & 17.0 & 19.0 \\
    & 200 & 27.5 & 19.5 & 0.5 & 3.0 & 22.0 & 18.0 & 21.5 \\
    & 500 & 28.0 & 29.0 & 0.0 & 1.5 & 30.0 & 21.0 & 24.0 \\
    & 1000 & 22.5 & 29.0 & 0.5 & 9.5 & 27.0 & 28.5 & 31.0 \\
    \midrule
    \textbf{10} 
    & 50 & 17.5 & 14.5 & 0.0 & 1.5 & 21.0 & 8.5 & 23.0 \\
    & 200 & 26.0 & 25.0 & 0.0 & 5.0 & 33.5 & 23.5 & 34.0 \\
    & 500 & 23.5 & 36.0 & 0.0 & 3.5 & 32.0 & 25.5 & 32.0 \\
    & 1000 & 32.5 & 39.0 & 0.0 & 5.0 & 38.0 & 33.5 & 38.5 \\
    \midrule
    \textbf{20} 
    & 50 & 21.5 & 16.5 & 0.0 & 1.5 & 30.0 & 16.0 & 29.0 \\
    & 200 & 24.5 & 20.0 & 0.0 & 5.5 & 35.5 & 34.5 & 33.0 \\
    & 500 & 29.5 & 34.0 & 0.0 & 6.0 & 41.0 & 43.5 & 27.5 \\
    & 1000 & 33.5 & 41.0 & 0.0 & 4.0 & 46.5 & 50.0 & 51.0 \\
    \midrule
    \multicolumn{2}{l}{\textbf{Average Rank (Lower is Better)}} & 3.62 & 3.52 & 6.57 & 6.00 & \underline{2.33} & 3.43 & \textbf{2.14} \\
    \bottomrule
  \end{tabular}
  \vspace{-0.15cm}
\end{table*}

\begin{table*}[h]
  \centering
  \caption{Success rates (\%) on LIBERO-goal suite}
  \label{tab:libero_goal_transposed}
  \begin{tabular}{lcccccccc}
    \toprule
    \textbf{{Demos}} & \textbf{{Steps}} & Octo & VQ-BeT & DP & DiT & \underline{H-Zero} & H-Gen & \textbf{\method} \\
    \midrule
    \textbf{1} & 0 & 0.0 & 0.0 & 0.0 & 0.0 & 0.0 & 2.5 & 1.5 \\
    & 50 & 10.5 & 16.0 & 0.0 & 3.0 & 38.5 & 21.0 & 38.5 \\
    & 200 & 22.0 & 35.0 & 0.0 & 5.5 & 43.5 & 34.5 & 38.0 \\
    & 500 & 33.0 & 43.0 & 1.0 & 15.0 & 44.5 & 35.5 & 41.5 \\
    & 1000 & 34.0 & 44.5 & 0.5 & 21.0 & 42.0 & 43.5 & 37.5 \\
    \midrule
    \textbf{3} & 50 & 12.5 & 20.0 & 0.5 & 3.0 & 47.5 & 29.0 & 52.5 \\
    & 200 & 14.5 & 39.0 & 1.0 & 5.5 & 45.5 & 35.5 & 66.5 \\
    & 500 & 20.5 & 58.0 & 0.5 & 12.0 & 52.0 & 46.5 & 67.5 \\
    & 1000 & 40.0 & 67.0 & 1.0 & 25.0 & 55.5 & 58.5 & 65.5 \\
    \midrule
    \textbf{5} & 50 & 16.0 & 15.5 & 0.5 & 4.0 & 49.0 & 50.0 & 60.0 \\
    & 200 & 38.0 & 48.0 & 1.5 & 2.5 & 54.5 & 44.5 & 70.0 \\
    & 500 & 45.5 & 56.0 & 0.5 & 9.5 & 59.0 & 66.0 & 73.0 \\
    & 1000 & 55.0 & 68.5 & 0.0 & 21.5 & 59.5 & 59.0 & 68.0 \\
    \midrule
    \textbf{10} & 50 & 16.5 & 20.5 & 0.0 & 2.0 & 63.0 & 45.0 & 75.5 \\
    & 200 & 40.5 & 52.0 & 0.5 & 4.5 & 61.0 & 60.5 & 74.0 \\
    & 500 & 46.0 & 72.5 & 1.0 & 6.5 & 68.0 & 77.5 & 65.0 \\
    & 1000 & 58.0 & 75.5 & 1.0 & 16.5 & 75.5 & 69.5 & 81.0 \\
    \midrule
    \textbf{20} & 50 & 13.5 & 23.5 & 0.5 & 1.0 & 52.0 & 43.0 & 68.0 \\
    & 200 & 41.0 & 57.0 & 2.0 & 4.5 & 73.0 & 61.0 & 76.5 \\
    & 500 & 62.0 & 71.5 & 1.5 & 6.5 & 79.5 & 69.5 & 82.5 \\
    & 1000 & 63.5 & 74.0 & 1.5 & 15.0 & 67.0 & 71.5 & 82.0 \\
    \midrule
    \multicolumn{2}{l}{\textbf{Average Rank (Lower is Better)}} & 4.86 & 2.90 & 6.81 & 5.86 & \underline{2.38} & 3.05 & \textbf{1.57} \\
    \bottomrule
  \end{tabular}
  \vspace{-0.15cm}
\end{table*}

\begin{table*}[h]
  \centering
  \caption{Success rates (\%) on LIBERO-object suite}
  \label{tab:libero_object_transposed}
  \begin{tabular}{lcccccccc}
    \toprule
    \textbf{{Demos}} & \textbf{{Steps}} & Octo & VQ-BeT & DP & DiT & H-Zero & H-Gen & \underline{\method} \\
    \midrule
    \textbf{1} & 0 & 0.0 & 0.0 & 0.0 & 0.0 & 0.0 & 0.0 & 0.0 \\
    & 50 & 0.0 & 0.0 & 0.0 & 0.0 & 5.0 & 9.5 & 3.5 \\
    & 200 & 1.0 & 2.5 & 0.0 & 0.0 & 2.5 & 9.0 & 1.5 \\
    & 500 & 0.0 & 8.5 & 0.0 & 0.0 & 5.0 & 8.5 & 0.0 \\
    & 1000 & 1.5 & 10.0 & 0.0 & 0.5 & 10.0 & 17.0 & 1.5 \\
    \midrule
    \textbf{3} & 50 & 0.0 & 0.0 & 0.0 & 0.0 & 4.0 & 0.0 & 1.5 \\
    & 200 & 1.5 & 28.0 & 0.0 & 0.0 & 3.5 & 5.0 & 15.0 \\
    & 500 & 5.5 & 18.0 & 0.0 & 0.0 & 13.0 & 8.5 & 22.5 \\
    & 1000 & 8.5 & 19.5 & 0.0 & 0.5 & 19.5 & 9.5 & 24.5 \\
    \midrule
    \textbf{5} & 50 & 0.0 & 4.5 & 0.0 & 0.0 & 5.0 & 9.0 & 15.0 \\
    & 200 & 5.0 & 27.5 & 0.0 & 0.0 & 3.5 & 5.5 & 12.0 \\
    & 500 & 4.0 & 25.5 & 0.0 & 0.0 & 24.5 & 13.5 & 22.0 \\
    & 1000 & 1.0 & 24.5 & 0.0 & 0.0 & 20.0 & 18.5 & 28.5 \\
    \midrule
    \textbf{10} & 50 & 0.0 & 4.5 & 0.0 & 0.0 & 21.5 & 22.5 & 5.0 \\
    & 200 & 6.0 & 35.5 & 0.0 & 0.0 & 24.5 & 2.5 & 8.0 \\
    & 500 & 16.0 & 16.5 & 0.0 & 0.0 & 31.5 & 14.5 & 33.5 \\
    & 1000 & 12.5 & 22.5 & 0.0 & 0.0 & 31.0 & 33.5 & 40.0 \\
    \midrule
    \textbf{20} & 50 & 0.0 & 2.5 & 0.0 & 0.0 & 9.0 & 16.5 & 17.5 \\
    & 200 & 4.0 & 45.0 & 0.0 & 0.0 & 18.0 & 35.0 & 20.0 \\
    & 500 & 13.5 & 31.0 & 0.0 & 0.0 & 26.0 & 52.0 & 30.0 \\
    & 1000 & 33.0 & 37.5 & 0.0 & 0.0 & 33.0 & 29.0 & 49.5 \\
    \midrule
    \multicolumn{2}{l}{\textbf{Average Rank (Lower is Better)}} & 4.95 & \underline{2.58} & 6.30 & 6.20 & 2.85 & 2.83 & \textbf{2.30} \\
    \bottomrule
  \end{tabular}
  \vspace{-0.15cm}
\end{table*}

\begin{table*}[h]
  \centering
  \caption{Success rates (\%) on LIBERO-spatial suite}
  \label{tab:libero_spatial_transposed}
  \begin{tabular}{lcccccccc}
    \toprule
    \textbf{{Demos}} & \textbf{{Steps}} & Octo & VQ-BeT & DP & DiT & \underline{H-Zero} & H-Gen & \textbf{\method} \\
    \midrule
    \textbf{1} & 0 & 0.0 & 3.0 & 0.0 & 0.0 & 0.0 & 0.0 & 0.0 \\
    & 50 & 27.5 & 11.5 & 0.0 & 0.0 & 30.5 & 23.0 & 31.0 \\
    & 200 & 29.5 & 26.5 & 0.0 & 2.0 & 34.5 & 38.5 & 45.5 \\
    & 500 & 36.0 & 36.5 & 0.5 & 8.0 & 37.0 & 42.5 & 46.5 \\
    & 1000 & 34.0 & 45.0 & 0.0 & 8.0 & 53.0 & 38.0 & 55.0 \\
    \midrule
    \textbf{3} & 50 & 22.5 & 13.5 & 0.5 & 0.0 & 34.0 & 36.5 & 48.5 \\
    & 200 & 44.5 & 36.5 & 0.0 & 1.0 & 43.0 & 45.0 & 54.5 \\
    & 500 & 45.5 & 50.5 & 0.0 & 4.0 & 49.5 & 53.5 & 68.0 \\
    & 1000 & 46.0 & 54.5 & 0.0 & 10.5 & 46.5 & 52.5 & 66.0 \\
    \midrule
    \textbf{5} & 50 & 41.0 & 25.0 & 1.0 & 0.0 & 43.5 & 41.5 & 62.5 \\
    & 200 & 54.0 & 44.5 & 0.0 & 0.5 & 57.5 & 53.0 & 58.5 \\
    & 500 & 59.0 & 47.5 & 0.0 & 4.5 & 50.0 & 70.0 & 60.5 \\
    & 1000 & 59.5 & 47.5 & 0.0 & 8.5 & 61.0 & 56.5 & 62.0 \\
    \midrule
    \textbf{10} & 50 & 43.5 & 23.5 & 1.0 & 0.0 & 48.5 & 55.5 & 53.5 \\
    & 200 & 62.0 & 45.5 & 0.0 & 0.0 & 50.5 & 48.5 & 68.5 \\
    & 500 & 57.5 & 49.5 & 0.0 & 3.0 & 59.5 & 58.5 & 70.5 \\
    & 1000 & 69.5 & 58.5 & 0.5 & 6.0 & 58.0 & 44.0 & 74.0 \\
    \midrule
    \textbf{20} & 50 & 43.0 & 26.5 & 1.0 & 0.0 & 49.5 & 46.0 & 60.0 \\
    & 200 & 63.5 & 47.5 & 0.0 & 0.0 & 61.5 & 60.0 & 67.5 \\
    & 500 & 69.5 & 67.5 & 0.0 & 1.5 & 57.0 & 56.5 & 76.0 \\
    & 1000 & 68.0 & 61.5 & 0.5 & 6.5 & 56.0 & 56.0 & 78.5 \\
    \midrule
    \multicolumn{2}{l}{\textbf{Average Rank (Lower is Better)}} & 3.38 & 4.14 & 6.43 & 6.00 & \underline{2.95} & 3.05 & \textbf{1.14} \\
    \bottomrule
  \end{tabular}
  \vspace{-0.15cm}
\end{table*}

\section{Few-Shot Adaptation Details}
\label{app:adaptation_details}

\subsection{Comparison with Parameter-Efficient Fine-Tuning}
To ensure a fair comparison, we control the number of trainable parameters to be approximately the same ($\approx$ 0.74M) across all methods. Specifically, for entangled architectures, we optimize LoRA parameters attached to the backbone; for hypernetwork-based architectures (including \method), we fine-tune the lightweight target policy generated by the frozen hypernetwork.

While traditional methods like Low-Rank Adaptation (LoRA)~\citep{hu2022lora} are effective for general adaptation, they face specific challenges when applied to policies with task-state entanglement. We compare the two approaches from the following perspectives:

\begin{itemize}
    \item \textbf{Decoupled vs. Entangled Updates:} LoRA updates must navigate through the backbone's entangled representations, where modifications for new language instructions can inadvertently affect established visuomotor features. In contrast, \method operates in a decoupled parameter space, exclusively updating the target policy $\theta_\pi$. This ensures that visual representation learning remains stable while allowing for rapid policy adaptation.
    
    \item \textbf{Initialization Quality:} Standard parameter-efficient methods typically initialize updates from scratch. Conversely, \method leverages the frozen hypernetwork $\mathcal{H}_\phi$ to provide a task-conditioned initialization. This places the optimization in a semantically informed region of the parameter space, significantly reducing the adaptation effort compared to learning offsets from a generic base model.
\end{itemize}

\subsection{Adaptation Algorithm Implementation}
For few-shot adaptation with $K$ demonstrations, we employ the procedure outlined in Algorithm~\ref{alg:adaptation}.

\begin{algorithm}[ht]
\caption{Few-Shot Adaptation for \method}
\label{alg:adaptation}
\begin{algorithmic}[1]
\Require Novel task instruction $l_{\text{new}}$, demonstrations $\mathcal{D}_{\text{new}} = \{\xi_j\}_{j=1}^K$, learning rate $\eta$
\Ensure Adapted policy parameters $\theta_\pi^*$
\State Generate initial parameters: $\theta_\pi^{(0)} \gets \mathcal{H}_\phi(\Phi_L(l_{\text{new}}))$
\State Initialize optimizer (e.g., Adam)
\For{step $t = 1$ to $T_{\text{max}}$}
    \State Sample batch $B$ from $\mathcal{D}_{\text{new}}$
    \State Compute loss: $\mathcal{L} = \frac{1}{|B|}\sum_{(o_k, a_k) \in B} \|\pi(o_k; \theta_\pi) - a_k\|_2^2$
    \State Update parameters: $\theta_\pi \gets \theta_\pi - \eta \nabla_{\theta_\pi}\mathcal{L}$
\EndFor
\State \Return $\theta_\pi^* \gets \theta_\pi$
\end{algorithmic}
\end{algorithm}

The key insight is that the hypernetwork-generated initialization $\theta_\pi^{(0)}$ provides a strong task-specific prior. This positions the optimization in a favorable region of the parameter space, enabling rapid convergence. The hypernetwork parameters $\phi$ remain frozen throughout this process, preserving the learned meta-knowledge. Typically, the adaptation converges within 50-200 gradient steps for $K \in [1, 5]$.

\subsection{More Few-shot Adaptation Results}
\label{app:more_results}
We present a comprehensive evaluation of few-shot adaptation performance across four LIBERO benchmark suites: LIBERO-10, LIBERO-Goal, LIBERO-Object, and LIBERO-Spatial. The detailed success rates across varying numbers of demonstrations ($K \in \{1, 3, 5, 10, 20\}$) and adaptation steps are reported in Tables~\ref{tab:libero_10_transposed}, \ref{tab:libero_goal_transposed}, \ref{tab:libero_object_transposed}, and \ref{tab:libero_spatial_transposed}.

\textbf{Overall Performance and Ranking.} 
As indicated by the average rank metrics, \method consistently outperforms baselines across all task suites. Specifically, \method achieves the best (lowest) average rank on LIBERO-10 (2.14), LIBERO-Goal (1.57), LIBERO-Spatial (1.14), and LIBERO-Object (2.30). This demonstrates the method's superior robustness and versatility across diverse manipulation tasks, regardless of whether the task requires long-horizon sequencing, precise spatial positioning, or object manipulation.

\textbf{Adaptation Efficiency.} 
A key advantage of \method is its rapid convergence. In the extremely low-data regime (e.g., $K=1$ or $K=3$), \method often achieves non-trivial success rates within just 50 to 200 gradient steps, whereas baselines like Octo typically require significantly more iterations or data to achieve a better performance. For instance, in LIBERO-Goal (Table~\ref{tab:libero_goal_transposed}) with only 1 demonstration, \method reaches 38.5\% success rate at 50 steps, significantly surpassing the baselines.

\textbf{Comparison with Baselines.} 
Previous strong baselines like VQ-BeT perform well in specific domains (e.g., object-centric tasks). However, \method surpasses VQ-BeT even on the LIBERO-Object suite (Avg Rank 2.30 vs. 2.58) while maintaining a significant lead in spatial and goal-conditioned tasks. Furthermore, \method consistently outperforms the ablation baseline H-Gen and H-Zero, validating the necessity and effectiveness of the two-stage parameter generation.

\section{Language Encoder Ablation}

We further investigate how the choice of language encoder affects the quality of the generated policy parameters. 
Table~\ref{tab:lang_encoder_ablation} reports the success rates on LIBERO-90 when using different off-the-shelf encoders to obtain the task embedding $e_l$.
\method achieves the highest performance when equipped with a T5-small encoder (94.3\%), while replacing T5-small with CLIP-base~\citep{radford2021learning} or BERT-base~\citep{devlin2019bert} leads to notable degradation (88.9\% and 81.2\%, respectively). 
We hypothesize that T5's text-to-text pre-training objective provides richer semantic and syntactic structure, enabling the hypernetwork to construct a more coherent parameter manifold from language input. 
In contrast, CLIP-base is primarily optimized for image-text alignment, and BERT-base focuses on masked-token prediction, both of which offer weaker task-level semantics for generating policy parameters.
Based on these results, we adopt T5-small as our default language encoder throughout all experiments.

\begin{table}[ht]
\centering
\caption{Ablation on the choice of language encoder.}
\begin{tabular}{l c c c}
\toprule
              & \textbf{T5-small} & \textbf{CLIP-base} & \textbf{BERT-base} \\ \midrule
Octo     & 84.7\%            & 85.7\%             & 79.3\%             \\
\textbf{\method} & 94.3\%            & 88.9\%             & 81.2\%             \\
\bottomrule
\end{tabular}
\label{tab:lang_encoder_ablation}
\end{table}

\section{Hypernetwork Architecture Ablation}

\begin{table}[t]
\centering
\caption{Ablation of hypernetwork components on LIBERO-90. Both coarse initialization (WIN) and iterative refinement are essential for performance.}
\label{tab:ablation-hypernet}
\begin{tabular}{lcccc}
\toprule
Method Variant & Easy & Medium & Long & Overall \\
\midrule
\method w/o WIN & 93.6 & 84.3 & 68.5 & 77.7 \\
\method w/ WIN only (no refinement) & 87.3 & 79.7 & 74.7 & 78.2 \\
\method (full) & \textbf{97.3} & \textbf{95.3} & \textbf{92.7} & \textbf{94.3} \\
\bottomrule
\end{tabular}
\vspace{-0.15cm}
\end{table}

To validate our two-stage, coarse-to-fine generation process, which combines a Weight Initialization Network (WIN) with iterative refinement, we ablate each component. The results in Table~\ref{tab:ablation-hypernet} demonstrate that both stages are critical for optimal performance. While removing the WIN module (\textbf{\method w/o WIN}) maintains high success on Easy tasks (93.6\%), performance drops significantly on Long-horizon tasks (68.5\% vs. 92.7\%), highlighting the difficulty of optimizing complex behaviors without a strong initialization. Conversely, using only the coarse initialization (\textbf{\method w/ WIN Only}) achieves a decent overall success rate of 78.2\% but lacks the precision required for harder tasks. Our \textbf{full model}, which integrates both stages, achieves the highest performance (94.3\%), substantially outperforming the WIN-only variant by +18.0\% on long-horizon tasks. This confirms that coarse initialization and iterative refinement are complementary, with their combination being essential for solving multi-stage manipulation tasks.

\section{Different Number of Tasks}

We vary the number of training tasks using subsets of LIBERO-90 and report success rates in Table~\ref{tab:task_diversity}. The performance trends for each method are shown below.

\begin{table}[ht]
\centering
\caption{Impact of Task Diversity (Number of Tasks) on LIBERO-90 Success Rate.}
\begin{tabular}{lccccc}
\toprule
\textbf{Num. Tasks} & \textbf{10} & \textbf{20} & \textbf{40} & \textbf{60} & \textbf{90} \\
\midrule
DP             & 88.0        & 87.3      & 76.8        & 73.6        & 75.2        \\
Octo           & 82.0        & 80.7        & 82.3        & 82.5        & 84.7        \\
\method           & 82.0        & 86.4        & 85.6        & 87.2        & 94.3        \\ \bottomrule
\end{tabular}
\label{tab:task_diversity}
\end{table}
From the results, we observe distinct scaling behaviors across the methods. Diffusion Policy (DP) suffers from negative scaling as task diversity increases: while it starts with a strong performance of 88.0\% on 10 tasks, its success rate drops significantly to 75.2\% as the number of tasks expands to 90, suggesting potential capacity limitations or task interference when handling a large variety of behaviors. Octo, on the other hand, displays relative stability with performance remaining largely flat (rising marginally from 82.0\% to 84.7\%), indicating that the architecture is robust but does not gain significant advantages from the increased training distribution. In contrast, \method demonstrates strong positive transfer; starting at 82.0\%, its performance improves consistently as more tasks are added, culminating in a 94.3\% success rate on the full 90-task suite. This scaling behavior also supports the hypothesis that \method’s hypernetwork learns a \textit{semantically structured parameter manifold}. Rather than being overwhelmed by task variety, our method leverages the richer task distribution to learn more generalizable policies, effectively outperforming task-state entangled architectures like Octo and Diffusion Policy in high-diversity settings.
\begin{figure*}[h]
    \centering
    \includegraphics[width=\textwidth]{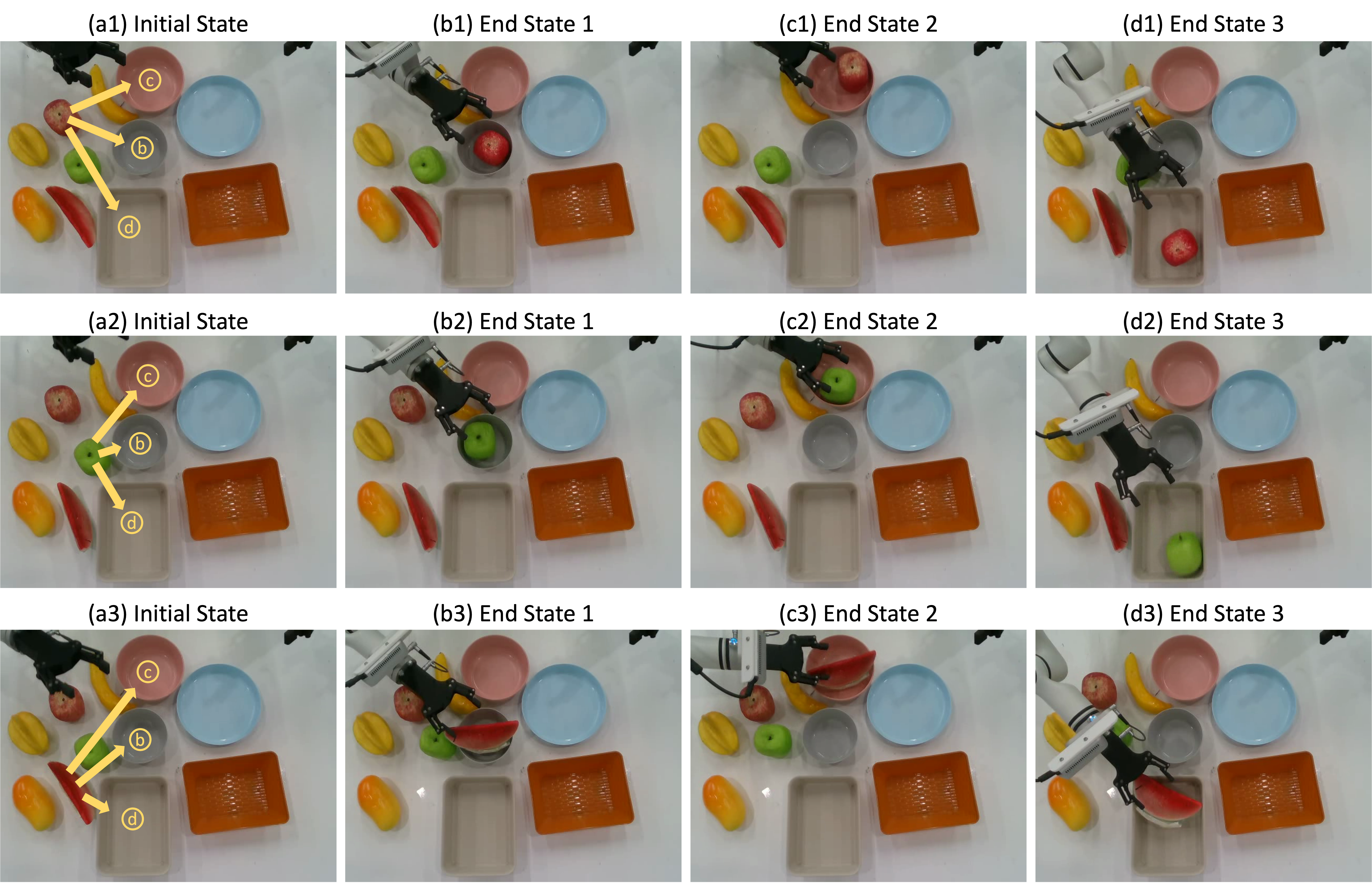}
    \caption{\textbf{Full Visualization of Real-World Combinatorial Tasks.}
    We evaluate \method on all 9 combinations of the task matrix. 
    Each row corresponds to a specific target object: \textit{Top Row:} Red Apple; \textit{Middle Row:} Green Apple; \textit{Bottom Row:} Watermelon Slice.
    \textit{Column 1 (Initial State):} Shows the cluttered starting configuration. Note that within each row, the visual observation is shared across all three downstream tasks.
    \textit{Columns 2-4 (Outcomes):} Show the successful execution of \method conditioned on three distinct instructions: 
    placing the object into the \textit{Small Gray Bowl} (Col 2), 
    the \textit{Large Red Bowl} (Col 3), 
    and the \textit{Light Gray Box} (Col 4).
    The yellow arrows in the initial states illustrate the three divergent trajectories the policy must generate based solely on language input.
    }
    \label{fig:app_real_matrix}
\end{figure*}

\section{Real Robot Experiments Details}
\label{app:real-robot}

In this section, we provide detailed specifications of the real-world experimental setup, the data collection protocol, and a comprehensive visualization of the combinatorial manipulation tasks discussed in the main paper.

\subsection{Hardware and Environment Setup}
The experimental setup features a RealMan 7-DoF robotic arm equipped with a parallel-jaw gripper. Visual observations are captured via two Intel RealSense D435i cameras: one mounted on a fixed pole to provide a static third-person view of the workspace, and another mounted on the end-effector to provide a gripper (egocentric) view. Both cameras stream RGB-D images at a resolution of $640 \times 480$, which are subsequently resized to $224 \times 224$ for processing.
The workspace contains a set of interactable objects (Red Apple, Green Apple, Watermelon Slice) and containers (Small Gray Bowl, Large Red Bowl, Light Gray Box), along with various distractor objects (e.g., bananas, mangoes, and other fruits) to increase visual clutter and occlusion.

\subsection{Task Definitions and Combinatorial Protocol}
To rigorously test language grounding, we constructed a combinatorial benchmark consisting of 9 distinct tasks. These tasks represent the Cartesian product of the object set $\mathcal{O}$ and the container set $\mathcal{C}$:
\begin{itemize}
    \item \textbf{Objects ($\mathcal{O}$):} \{Red Apple, Green Apple, Watermelon Slice\}
    \item \textbf{Containers ($\mathcal{C}$):} \{Small Gray Bowl, Large Red Bowl, Light Gray Box\}
\end{itemize}
This results in $3 \times 3 = 9$ unique language instructions of the form ``Pick up the [Object] and place it into the [Container].''

\textbf{Visual Ambiguity Control.} A critical feature of this benchmark is the preservation of visual context. 
As shown in Figure~\ref{fig:app_real_matrix}, the workspace maintains a globally consistent, cluttered visual context with all candidate objects simultaneously present.
Consequently, the robot must rely solely on the linguistic instruction to determine whether to transport the object to destination (b), (c), or (d). This design explicitly penalizes methods that rely on observation leakage or memorize scene-to-action shortcuts.

\subsection{Data Collection and Evaluation}
\paragraph{Data Collection}
We collected expert demonstrations using a teleoperation interface. For each of the 9 tasks, we recorded 100 successful episodes, resulting in a total dataset of 900 episodes.

\paragraph{Evaluation Metrics}
During evaluation, a trial is considered a \textit{success} if and only if the following criteria are met:
\begin{enumerate}
    \item The robot successfully grasps the correct target object specified in the instruction.
    \item The object is transported to and released inside the correct target container.
\end{enumerate}
We evaluate each method over 30 trials per task, totaling $30 \times 9 = 270$ evaluation episodes per method.

\begin{figure}[t]
    \centering
    \includegraphics[width=\linewidth]{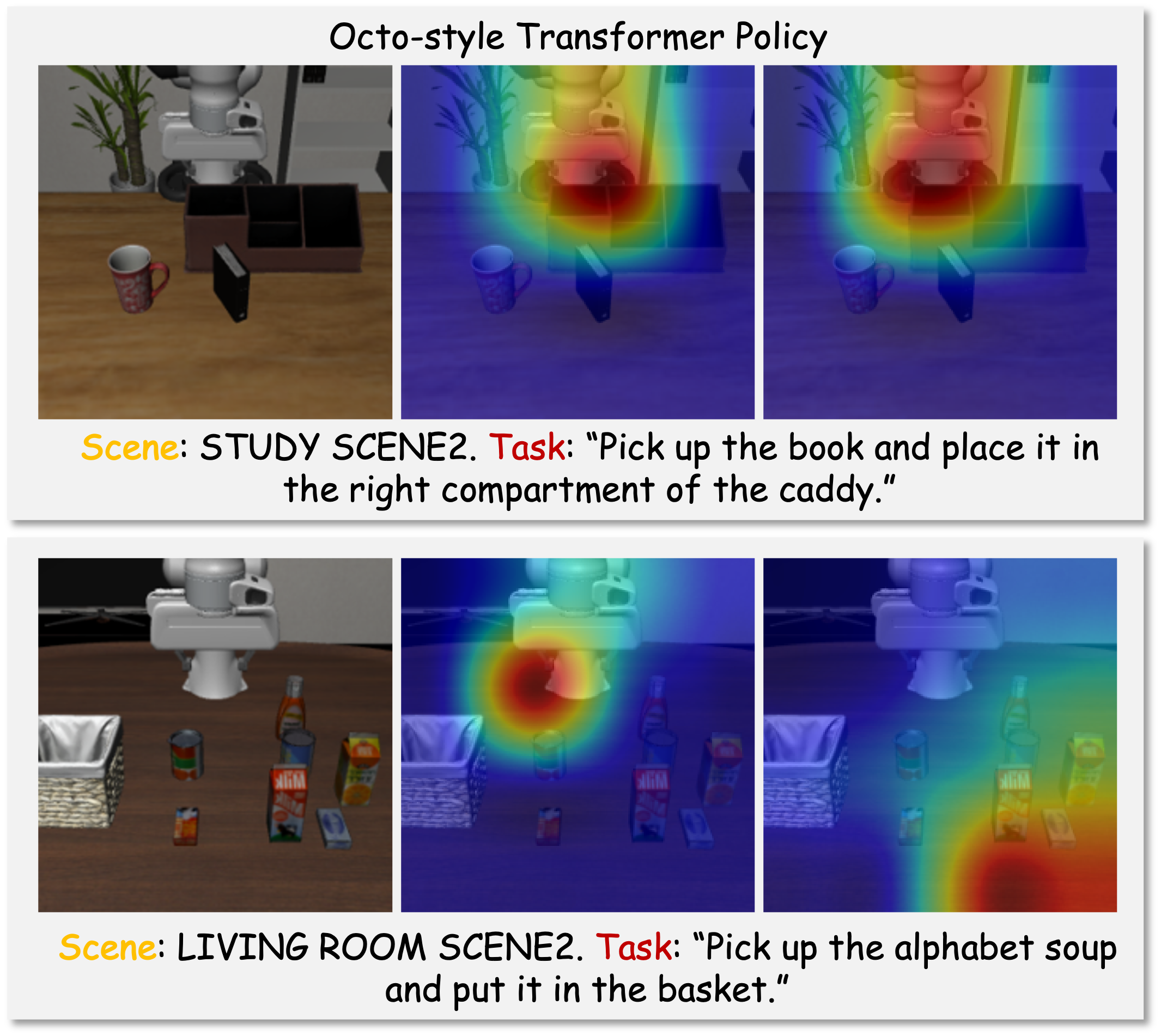}
    \caption{
        \textbf{Attention Map Visualization for Task-State Entangled Architectures.} 
        We visualize attention maps for an Octo-style Transformer policy on two evaluation tasks.
        Despite specific language instructions (e.g., ``Pick up the book''), the attention mechanisms frequently fail to localize the task-relevant objects. 
        Instead, attention is heavily concentrated on the robot's manipulator arm (a constant visual feature) or irrelevant background objects. 
        This provides visual evidence of \textit{observation leakage}: the policy relies on scene-to-action shortcuts rather than true language grounding.
    }
    \label{fig:attention_maps_apx}
\end{figure}

\section{Extended Qualitative Analysis}
\label{app:visualization}

To provide concrete evidence for our claims regarding the limitations of task-state entanglement, we conduct qualitative analyses on a representative baseline. These visualizations empirically demonstrate the phenomena of attention misalignment and observation leakage, validating our hypothesis that entangled architectures often fail to ground language instructions in the visual scene.

\subsection{Visual Evidence of Observation Leakage}
We visualize the attention maps from the Transformer layers of a representative entangled baseline: an Octo-style Transformer Policy. As illustrated in Figure~\ref{fig:attention_maps_apx}, these visualizations reveal a significant disconnect between the semantic requirements of the instruction and the model's actual visual focus.

For instance, in the top row, the instruction explicitly specifies ``Pick up the \textbf{book}...''. However, the policy concentrates its attention almost exclusively on the robot's gripper and the unrelated caddy structure, ignoring the target object entirely. Similarly, in the bottom row (``Pick up the \textbf{alphabet soup}...''), the policy fails to focus on the target object. Instead, its attention maps highlight the gripper or diffuse background regions.

This pattern consistently demonstrates the mechanism of observation leakage: in the absence of explicit structural decoupling, the entangled network learns to rely on spurious correlations, such as the ever-present robot arm, rather than grounding the variable linguistic concepts (e.g., ``book,'' ``soup''). The model ``cheats'' by mapping the constant visual feature (the gripper) to actions, bypassing the language instruction entirely.

In contrast, \method blocks this direct shortcut pathway by construction. Since the target policy receives no language input, task identity must be encoded into the generated policy weights by the hypernetwork, enforcing a structural dependency on instruction semantics rather than visual shortcuts.

\section{Detailed Task Splits}
\label{app:full_task_splits}

In this section, we provide the complete list of tasks used in our experiments for both LIBERO-90 and Meta-World. 

\subsection{LIBERO Task Splits}
\label{app:libero-splits}

90 tasks from LIBERO-90 are divided as follows:

\noindent\textbf{Training Tasks (90):}
\begin{itemize}[topsep=2pt, itemsep=1pt]
    \item \textbf{Easy (11 tasks):} Single-object manipulation (e.g., ``turn on the stove'')
    \item \textbf{Medium (35 tasks):} Multi-object coordination (e.g., ``put the black bowl on the plate'')  
    \item \textbf{Long (44 tasks):} Sequential multi-step (e.g., ``turn on the stove and put the frying pan on it'')
\end{itemize}
The exact split of tasks is listed in Table~\ref{tab:libero-tasks-part1} and~\ref{tab:libero-tasks-part2}.

\subsection{Meta-World Task Splits}
\label{app:metaworld-splits}
Similarly, we split Meta-World ML45 into 4 categories: 

\noindent\textbf{Training Tasks (45):}
\begin{itemize}
    \item \textbf{open\&close (8 tasks)}: Open or close an object (e.g., ``window open'')
    \item \textbf{pick\&place (4 tasks)}: Pick the object up or place it elsewhere (e.g., ``pick out of hole'')
    \item \textbf{press\&pull (16 tasks)}: Press the button or lever (e.g., ``button press topdown'')
    \item \textbf{others (17 tasks)}: Other tasks not in previous categories (e.g., ``soccer'')
\end{itemize}

These tasks correspond to the 45 training environments in the standard Meta-World ML45 split. The exact split of tasks is listed in Table~\ref{tab:metaworld-tasks}.

\begin{table*}[t]
\centering
\caption{LIBERO-90 Training Tasks (Part 1: One-Object \& Two-Object Short)}
\label{tab:libero-tasks-part1}
\begin{tabular}{p{0.15\linewidth}p{0.8\linewidth}}
\toprule
\textbf{Category} & \textbf{Tasks} \\
\midrule
\multirow{11}{*}{One-Object} 
& KITCHEN\_SCENE3\_turn\_on\_the\_stove \\
& KITCHEN\_SCENE4\_close\_the\_bottom\_drawer\_of\_the\_cabinet \\
& KITCHEN\_SCENE8\_turn\_off\_the\_stove \\
& KITCHEN\_SCENE9\_turn\_on\_the\_stove \\
& KITCHEN\_SCENE10\_close\_the\_top\_drawer\_of\_the\_cabinet \\
& KITCHEN\_SCENE2\_open\_the\_top\_drawer\_of\_the\_cabinet \\
& KITCHEN\_SCENE5\_close\_the\_top\_drawer\_of\_the\_cabinet \\
& KITCHEN\_SCENE7\_open\_the\_microwave \\
& KITCHEN\_SCENE1\_open\_the\_top\_drawer\_of\_the\_cabinet \\
& KITCHEN\_SCENE1\_open\_the\_bottom\_drawer\_of\_the\_cabinet \\
& KITCHEN\_SCENE6\_close\_the\_microwave \\
\midrule
\multirow{35}{*}{Two-Object}
& KITCHEN\_SCENE3\_put\_the\_frying\_pan\_on\_the\_stove \\
& KITCHEN\_SCENE3\_put\_the\_moka\_pot\_on\_the\_stove \\
& KITCHEN\_SCENE4\_put\_the\_wine\_bottle\_on\_the\_wine\_rack \\
& KITCHEN\_SCENE4\_put\_the\_black\_bowl\_on\_top\_of\_the\_cabinet \\
& KITCHEN\_SCENE4\_put\_the\_black\_bowl\_in\_the\_bottom\_drawer\_of\_the\_cabinet \\
& KITCHEN\_SCENE6\_put\_the\_yellow\_and\_white\_mug\_to\_the\_front\_of\_the\_white\_mug \\
& KITCHEN\_SCENE8\_put\_the\_right\_moka\_pot\_on\_the\_stove \\
& KITCHEN\_SCENE9\_put\_the\_frying\_pan\_on\_top\_of\_the\_cabinet \\
& KITCHEN\_SCENE9\_put\_the\_white\_bowl\_on\_top\_of\_the\_cabinet \\
& KITCHEN\_SCENE9\_put\_the\_frying\_pan\_on\_the\_cabinet\_shelf \\
& KITCHEN\_SCENE9\_put\_the\_frying\_pan\_under\_the\_cabinet\_shelf \\
& KITCHEN\_SCENE10\_put\_the\_black\_bowl\_in\_the\_top\_drawer\_of\_the\_cabinet \\
& LIVING\_ROOM\_SCENE5\_put\_the\_white\_mug\_on\_the\_left\_plate \\
& LIVING\_ROOM\_SCENE5\_put\_the\_yellow\_and\_white\_mug\_on\_the\_right\_plate \\
& LIVING\_ROOM\_SCENE5\_put\_the\_red\_mug\_on\_the\_left\_plate \\
& LIVING\_ROOM\_SCENE5\_put\_the\_red\_mug\_on\_the\_right\_plate \\
& LIVING\_ROOM\_SCENE6\_put\_the\_white\_mug\_on\_the\_plate \\
& LIVING\_ROOM\_SCENE6\_put\_the\_red\_mug\_on\_the\_plate \\
& LIVING\_ROOM\_SCENE6\_put\_the\_chocolate\_pudding\_to\_the\_left\_of\_the\_plate \\
& LIVING\_ROOM\_SCENE6\_put\_the\_chocolate\_pudding\_to\_the\_right\_of\_the\_plate \\
& KITCHEN\_SCENE2\_put\_the\_black\_bowl\_at\_the\_back\_on\_the\_plate \\
& KITCHEN\_SCENE2\_put\_the\_black\_bowl\_at\_the\_front\_on\_the\_plate \\
& KITCHEN\_SCENE2\_put\_the\_middle\_black\_bowl\_on\_the\_plate \\
& KITCHEN\_SCENE2\_put\_the\_middle\_black\_bowl\_on\_top\_of\_the\_cabinet \\
& KITCHEN\_SCENE2\_stack\_the\_middle\_black\_bowl\_on\_the\_back\_black\_bowl \\
& KITCHEN\_SCENE2\_stack\_the\_black\_bowl\_at\_the\_front\_on\_the\_black\_bowl\_in\_the\_middle \\
& KITCHEN\_SCENE5\_put\_the\_black\_bowl\_in\_the\_top\_drawer\_of\_the\_cabinet \\
& KITCHEN\_SCENE5\_put\_the\_black\_bowl\_on\_the\_plate \\
& KITCHEN\_SCENE5\_put\_the\_ketchup\_in\_the\_top\_drawer\_of\_the\_cabinet \\
& KITCHEN\_SCENE5\_put\_the\_black\_bowl\_on\_top\_of\_the\_cabinet \\
& KITCHEN\_SCENE7\_put\_the\_white\_bowl\_on\_the\_plate \\
& KITCHEN\_SCENE7\_put\_the\_white\_bowl\_to\_the\_right\_of\_the\_plate \\
& KITCHEN\_SCENE1\_put\_the\_black\_bowl\_on\_top\_of\_the\_cabinet \\
& KITCHEN\_SCENE1\_put\_the\_black\_bowl\_on\_the\_plate \\
& KITCHEN\_SCENE4\_put\_the\_wine\_bottle\_in\_the\_bottom\_drawer\_of\_the\_cabinet \\
\bottomrule
\end{tabular}
\end{table*}

\begin{table*}[t]
\centering
\caption{LIBERO-90 Training Tasks (Part 2: Long Horizon)}
\label{tab:libero-tasks-part2}
\begin{tabular}{p{0.15\linewidth}p{0.8\linewidth}}
\toprule
\textbf{Category} & \textbf{Tasks} \\
\midrule
\multirow{44}{*}{Long-Horizon}
& KITCHEN\_SCENE4\_close\_the\_bottom\_drawer\_of\_the\_cabinet\_and\_open\_the\_top\_drawer \\
& KITCHEN\_SCENE9\_turn\_on\_the\_stove\_and\_put\_the\_frying\_pan\_on\_it \\
& KITCHEN\_SCENE10\_close\_the\_top\_drawer\_of\_the\_cabinet\_and\_put\_the\_black\_bowl\_on\_top\_of\_it \\
& KITCHEN\_SCENE10\_put\_the\_butter\_at\_the\_front\_in\_the\_top\_drawer\_of\_the\_cabinet\_and\_close\_it \\
& KITCHEN\_SCENE10\_put\_the\_butter\_at\_the\_back\_in\_the\_top\_drawer\_of\_the\_cabinet\_and\_close\_it \\
& KITCHEN\_SCENE10\_put\_the\_chocolate\_pudding\_in\_the\_top\_drawer\_of\_the\_cabinet\_and\_close\_it \\
& LIVING\_ROOM\_SCENE1\_pick\_up\_the\_cream\_cheese\_box\_and\_put\_it\_in\_the\_basket \\
& LIVING\_ROOM\_SCENE1\_pick\_up\_the\_alphabet\_soup\_and\_put\_it\_in\_the\_basket \\
& LIVING\_ROOM\_SCENE1\_pick\_up\_the\_tomato\_sauce\_and\_put\_it\_in\_the\_basket \\
& LIVING\_ROOM\_SCENE2\_pick\_up\_the\_orange\_juice\_and\_put\_it\_in\_the\_basket \\
& LIVING\_ROOM\_SCENE2\_pick\_up\_the\_milk\_and\_put\_it\_in\_the\_basket \\
& LIVING\_ROOM\_SCENE2\_pick\_up\_the\_butter\_and\_put\_it\_in\_the\_basket \\
& LIVING\_ROOM\_SCENE2\_pick\_up\_the\_alphabet\_soup\_and\_put\_it\_in\_the\_basket \\
& LIVING\_ROOM\_SCENE2\_pick\_up\_the\_tomato\_sauce\_and\_put\_it\_in\_the\_basket \\
& LIVING\_ROOM\_SCENE4\_stack\_the\_left\_bowl\_on\_the\_right\_bowl\_and\_place\_them\_in\_the\_tray \\
& LIVING\_ROOM\_SCENE4\_stack\_the\_right\_bowl\_on\_the\_left\_bowl\_and\_place\_them\_in\_the\_tray \\
& LIVING\_ROOM\_SCENE4\_pick\_up\_the\_black\_bowl\_on\_the\_left\_and\_put\_it\_in\_the\_tray \\
& LIVING\_ROOM\_SCENE4\_pick\_up\_the\_salad\_dressing\_and\_put\_it\_in\_the\_tray \\
& LIVING\_ROOM\_SCENE4\_pick\_up\_the\_chocolate\_pudding\_and\_put\_it\_in\_the\_tray \\
& STUDY\_SCENE1\_pick\_up\_the\_book\_and\_place\_it\_in\_the\_right\_compartment\_of\_the\_caddy \\
& STUDY\_SCENE1\_pick\_up\_the\_book\_and\_place\_it\_in\_the\_front\_compartment\_of\_the\_caddy \\
& STUDY\_SCENE1\_pick\_up\_the\_yellow\_and\_white\_mug\_and\_place\_it\_to\_the\_right\_of\_the\_caddy \\
& STUDY\_SCENE2\_pick\_up\_the\_book\_and\_place\_it\_in\_the\_left\_compartment\_of\_the\_caddy \\
& STUDY\_SCENE2\_pick\_up\_the\_book\_and\_place\_it\_in\_the\_right\_compartment\_of\_the\_caddy \\
& STUDY\_SCENE2\_pick\_up\_the\_book\_and\_place\_it\_in\_the\_back\_compartment\_of\_the\_caddy \\
& STUDY\_SCENE2\_pick\_up\_the\_book\_and\_place\_it\_in\_the\_front\_compartment\_of\_the\_caddy \\
& STUDY\_SCENE3\_pick\_up\_the\_book\_and\_place\_it\_in\_the\_left\_compartment\_of\_the\_caddy \\
& STUDY\_SCENE3\_pick\_up\_the\_book\_and\_place\_it\_in\_the\_right\_compartment\_of\_the\_caddy \\
& STUDY\_SCENE3\_pick\_up\_the\_book\_and\_place\_it\_in\_the\_front\_compartment\_of\_the\_caddy \\
& STUDY\_SCENE3\_pick\_up\_the\_white\_mug\_and\_place\_it\_to\_the\_right\_of\_the\_caddy \\
& STUDY\_SCENE3\_pick\_up\_the\_red\_mug\_and\_place\_it\_to\_the\_right\_of\_the\_caddy \\
& STUDY\_SCENE4\_pick\_up\_the\_book\_on\_the\_right\_and\_place\_it\_under\_the\_cabinet\_shelf \\
& STUDY\_SCENE4\_pick\_up\_the\_book\_in\_the\_middle\_and\_place\_it\_on\_the\_cabinet\_shelf \\
& STUDY\_SCENE4\_pick\_up\_the\_book\_on\_the\_left\_and\_place\_it\_on\_top\_of\_the\_shelf \\
& STUDY\_SCENE4\_pick\_up\_the\_book\_on\_the\_right\_and\_place\_it\_on\_the\_cabinet\_shelf \\
& LIVING\_ROOM\_SCENE3\_pick\_up\_the\_alphabet\_soup\_and\_put\_it\_in\_the\_tray \\
& LIVING\_ROOM\_SCENE3\_pick\_up\_the\_tomato\_sauce\_and\_put\_it\_in\_the\_tray \\
& LIVING\_ROOM\_SCENE3\_pick\_up\_the\_butter\_and\_put\_it\_in\_the\_tray \\
& LIVING\_ROOM\_SCENE3\_pick\_up\_the\_cream\_cheese\_and\_put\_it\_in\_the\_tray \\
& LIVING\_ROOM\_SCENE3\_pick\_up\_the\_ketchup\_and\_put\_it\_in\_the\_tray \\
& KITCHEN\_SCENE1\_open\_the\_top\_drawer\_of\_the\_cabinet\_and\_put\_the\_bowl\_in\_it \\
& LIVING\_ROOM\_SCENE1\_pick\_up\_the\_ketchup\_and\_put\_it\_in\_the\_basket \\
& STUDY\_SCENE1\_pick\_up\_the\_book\_and\_place\_it\_in\_the\_left\_compartment\_of\_the\_caddy \\
& KITCHEN\_SCENE3\_turn\_on\_the\_stove\_and\_put\_the\_frying\_pan\_on\_it \\
\bottomrule
\end{tabular}
\end{table*}

\begin{table*}[t]
    \centering
    \caption{Meta-World ML45 Training Task List}
    \label{tab:metaworld-tasks}
    \begin{tabular}{p{0.15\linewidth}p{0.8\linewidth}}
    \toprule
    \textbf{Category} & \textbf{Tasks} \\
    \midrule
    Pick \& Place 
    & pick-place, pick-place-wall, pick-out-of-hole, shelf-place, bin-picking \\
    \midrule
    Push/Press
    & push, stick-push, coffee-push, push-wall, handle-press-side, handle-press, button-press-wall, button-press-topdown, button-press-topdown-wall, button-press, lever-pull, stick-pull, handle-pull-side, handle-pull, coffee-pull, push-back \\
    \midrule
    Open \& Close
    & faucet-open, faucet-close, window-close, window-open, door-open, door-close, drawer-close, drawer-open, box-close \\
    \midrule
    Others
    & sweep, assembly, dial-turn, coffee-button, basketball, sweep-into, disassemble, hammer, plate-slide, plate-slide-side, soccer, plate-slide-back-side, plate-slide-back, reach, reach-wall, peg-insert-side, peg-unplug-side, hand-insert, door-lock, door-unlock \\
    \bottomrule
    \end{tabular}
\end{table*}

\clearpage

\end{document}